\documentclass[conference]{IEEEtran}
\IEEEoverridecommandlockouts
\usepackage{cite}
\usepackage{balance}
\usepackage{amsmath,amssymb,amsfonts}
\usepackage{algorithmic}
\usepackage{graphicx}
\usepackage{textcomp}
\usepackage{xcolor}
\usepackage{color}
\usepackage{colortbl}
\usepackage{multirow}
\usepackage{amsthm}
\usepackage{bm}
\usepackage{tabularx}
\usepackage{bbm}
\usepackage{subcaption}
\usepackage{subcaption}
\usepackage{caption}
\usepackage{threeparttable}
\usepackage[justification=centering]{caption}
\usepackage{xspace}
\usepackage{weiwAlgorithm}
\usepackage{paralist}
\usepackage{hyperref}
\usepackage{nccmath}

\theoremstyle{definition}
\newtheorem{definition}{Definition}[section]

\SetKwInput{KwInput}{Input}                
\SetKwInput{KwOutput}{Output}
\SetKwFunction{Enum}{Enumerate}

\def\BibTeX{{\rm B\kern-.05em{\sc i\kern-.025em b}\kern-.08em
    T\kern-.1667em\lower.7ex\hbox{E}\kern-.125emX}}

\renewcommand{\baselinestretch}{0.936}

\DeclareMathOperator*{\argmax}{arg\,max}

\begin{document}

\title{Reinforcement Learning Based Query Vertex Ordering Model for Subgraph Matching
}

\author{\IEEEauthorblockN{Hanchen Wang$^{\flat \ddagger}$, Ying Zhang$^{\flat \dagger}$\textsuperscript{*}, Lu Qin$^{\dagger}$, Wei Wang$^{\natural}$, Wenjie Zhang$^\ddagger$, Xuemin Lin$^\ddagger$}
\IEEEauthorblockA{$^{\flat}$Zhejiang Gongshang University,$^\ddagger$University of New South Wales,\\
$^{\dagger}$University of Technology Sydney,
$^{\natural}$Hong Kong University of Science and Technology\\
\{hanchen.wang, ying.zhang, lu.qin\}@uts.edu.au, weiwcs@ust.hk,  \{zhangw, lxue\}@cse.unsw.edu.au}
}

\maketitle

\begingroup\renewcommand\thefootnote{*}
\footnotetext{Corresponding Author}

\begin{abstract}
Subgraph matching is a fundamental problem in various fields that use graph structured data.
Subgraph matching algorithms enumerate all isomorphic embeddings of a query graph $q$ in a data graph $G$.
An important branch of matching algorithms exploit the backtracking search approach which recursively extends intermediate results following a matching order of query vertices.
It has been shown that the matching order plays a critical role in time efficiency of these backtracking based subgraph matching algorithms.
In recent years, many advanced techniques for query vertex ordering (i.e., matching order generation) have been proposed to reduce the unpromising intermediate results according to the preset heuristic rules.
In this paper, for the first time we apply the Reinforcement Learning (RL) and Graph Neural Networks (GNNs)
techniques to generate the high-quality matching order for subgraph matching algorithms.
Instead of using the fixed heuristics to generate the matching order, our model could capture and make full use of the graph information, and thus determine the query vertex order with the adaptive learning-based rule that could significantly 
reduces the number of redundant enumerations.
With the help of the reinforcement learning framework, our model is able to consider the long-term benefits rather than only consider the local information at current ordering step.
Extensive experiments on six real-life data graphs demonstrate that our proposed matching order generation technique
could reduce up to two orders of magnitude of query processing time compared to the state-of-the-art algorithms.
\end{abstract}

\begin{IEEEkeywords}
Subgraph Matching, Reinforcement Learning, Graph Neural Network
\end{IEEEkeywords}

\newcommand\mname{RL-QVO\xspace}
\newcommand\hname{Hybrid\xspace}

\section{Introduction}

In recent years, graph structured data has been increasingly ubiquitous.
Graph analysis also becomes much more important and popular in the area of data analytics.
Subgraph matching is one of the most fundamental problems in graph analysis.
Subgraph matching aims to find all embeddings in a data graph $G$ that are isomorphic to a query graph $q$.
For example, in Figure~\ref{fig:intro}, $\{(u_1, v_1), (u_2, v_4), (u_3, v_5),$ $(u_4, v_{10})\}$ is a match from query graph $q$ to data graph $G$.
Due to its importance, subgraph matching is widely used in various areas in both academia and industry~\cite{DBLP:journals/pvldb/SahuMSLO17,DBLP:journals/bioinformatics/PrzuljCJ06,snijders2006new,yan2004graph}.

Subgraph matching has been proved to be a NP-complete problem~\cite{DBLP:books/fm/GareyJ79}, which means that the computational complexity is factorial in the worst case.
Though this limit is intrinsic and cannot be overcome, great research efforts still lead to significant advances in reducing unpromising intermediate results, and thus reduce the computational complexity in the average case.

\begin{figure}[tb]
    \centering
    \begin{subfigure}[t]{0.132\textwidth}
    \centering
    \includegraphics[width=\textwidth]{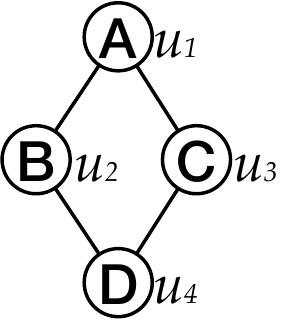}
    \caption{Query Graph $q$}
    \label{fig:query_graph} %
    \end{subfigure}
    \begin{subfigure}[t]{0.345\textwidth}
    \centering
    \includegraphics[width=\textwidth]{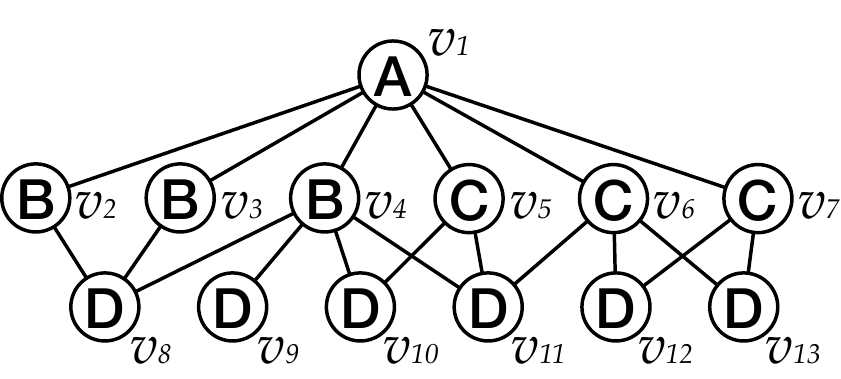}
    \caption{Data Graph $G$}
    \label{fig:data_graph} %
    \end{subfigure}
    \vspace{-1mm}
    \caption{Example query graph and data graph.}
    \vspace{-6mm}
    \label{fig:intro}
\end{figure}

The subgraph matching algorithms can be categorized in two major classes by their main methodologies: join-based algorithms and backtracking search based algorithms.
The join-based approaches~\cite{DBLP:journals/pvldb/LaiQLC15,DBLP:journals/pvldb/LaiQLZC16,DBLP:journals/pvldb/LaiQYJLWHLQZZQZ19} convert the query $q$ to a multi-way join by mapping vertices and edges in $q$ to attributes and relations. 
These approaches evaluate the multi-way join to find matching results.
Another class of approaches are the backtracking search based algorithms, which recursively extend intermediate results by mapping query vertices to data vertices along an order of query vertices~\cite{DBLP:journals/pvldb/SunWWSL12}.
Many methods adopt the state space representation for subgraph matching, where each state represents an intermediate result~\cite{carletti2017challenging}, \textit{i.e.,} a mapping from query vertex to data vertex.
The methods explore the state space with given order to find feasible state as a match.
Among these methods, the query vertex order plays an important role in reducing the average time complexity for subgraph matching, because a well-designed query vertex order could enable the algorithm to prune out the unpromising results as much as possible~\cite{DBLP:conf/sigmod/BiCLQZ16,DBLP:conf/sigmod/HanKGPH19,han2013turboiso}.

In this paper, we focus on the backtracking search based subgraph matching algorithms. 
As shown in~\cite{sun2020memory}, it is critical to choose a good matching order for the backtracking search based algorithms because it will greatly affect the search performance. 
Therefore, existing backtracking based algorithms put a lot of efforts to generate good matching orders.
Clearly, finding the optimal matching order is time-prohibitive, and existing algorithms resort to a variety of heuristic strategies. 
For instance, QuickSI~\cite{shang2008taming} proposed a \textit{infrequent-edge first} ordering method which converts $q$ into a weighted graph based on the label frequency of query edges in data graph and generates the matching order along the ascending order of their weights.
RI~\cite{bonnici2013subgraph} generates the order based on the structure of $q$, which selects the vertex with the maximum degree in $q$ as the first vertices, and then picks the vertex that has the most neighbors in order $\phi$ as the next node.
Other methods such as VF2++~\cite{juttner2018vf2++}, GraphQL~\cite{DBLP:conf/sigmod/HeS08}, CFL~\cite{DBLP:conf/sigmod/BiCLQZ16} CECI~\cite{DBLP:conf/sigmod/BhattaraiLH19},  DP-iso~\cite{DBLP:conf/sigmod/HanKGPH19} and VEQ~\cite{kim2021versatile} generate the order $\phi$ by building auxiliary structure of $q$ based on the degrees, labels and candidate sizes of query vertices.

However, these ordering methods only utilize the local structural and label information by greedy heuristics.
Consequently, these methods are lacking the ability of fully exploiting the information within the query and data graphs; meanwhile, these methods can only obtain the locally optimal order produced by the greedy search algorithm. 
The auxiliary structure-based query ordering methods in CFL and DP-iso, which dynamically generate matching orders, can result in a large number of unsolved queries (\textit{i.e.,} whose query processing time exceeds the preset time limit), and hence cannot achieve the state-of-the-art search performance according to the empirical study~\cite{sun2020memory}.


\vspace{1mm}
\noindent \textbf{Motivations and Our Approaches}

In this paper, we propose the first \underline{R}einforcement \underline{L}earning based \underline{Q}uery \underline{V}ertex \underline{O}rdering method for
backtracking based subgraph matching, namely \mname, to fill this research gap.
\mname\ aims to be a more efficient and scalable technique to conduct subgraph matching.
We introduce the motivations in this paper as follows:

\vspace{1mm}
\noindent \textbf{Motivation 1: Make Full Use of the Graph Information.}
As mentioned previously, the recent subgraph matching models generate $\phi$ based on the preset priority of heuristics such as degree, label frequency, path frequency, candidate size and etc.
In practice, the order is usually determined by partial heuristics, and other statistics are somehow ignored.
Besides, the priority of the heuristics cannot be adapted for graphs with different distributions.
Consider $q$ and $G$ in Figure~\ref{fig:intro}, if we use the ordering method of RI which selects the first node with the highest degree, the order will be generated randomly due to the structural symmetry of $q$.
Accordingly, the following order is also generated on a random basis.
Clearly, the order generated by RI cannot guarantee to reduce the redundant intermediate results for this query.
Otherwise, if we build the order according to the label frequency in this case, vertex $v_1$ will be selected as the starting vertex in $\phi$, which eventually reduces unpromising intermediate results.
Therefore, adaptive priority for heuristics under particular query graph could significantly enhance the ordering methods, which requires the whole picture of graph information and decision making based on these information.

\vspace{1mm}
\noindent \textit{Our Approach: Capture the Graph Information with Graph Neural Network.}
Regarding the above observation, we exploit the graph neural networks~\cite{DBLP:conf/iclr/KipfW17,DBLP:conf/iclr/VelickovicCCRLB18,wu2020comprehensive} whose ability of feature extraction and representation learning has been widely proven theoretically and empirically, to capture the information hidden in $q$.
We carefully design the initial feature for each query vertex based on the popular heuristics.
With the help of GNN's message passing for neighbor aggregation, \mname\ can naturally capture the query graph's structural and attribute information.
Furthermore, thanks to the generality of the machine learning based techniques, \mname\ is able to select the next node to generate the matching order for different query graphs according to learned vertex representations.

\vspace{1mm}
\noindent \textbf{Motivation 2: Limitation of Greedy Heuristics.}
Most existing ordering methods utilize the greedy search methods with preset rules which will intrinsically fall into local optimum.
Such methods intend to add the vertex that could reduce the most redundant intermediate results to the order $\phi$ at every step.
However, these methods cannot consider the long-term query time cost (\textit{i.e.,} quality of the order).
The exact optimal order can only be found after all possible order permutations are evaluated.
Therefore, it is challenging for us to develop a subgraph matching algorithm that can overcome the limitation of the greedy criterion.

\vspace{1mm}
\noindent \textit{Our Approach: Overcoming the Limitation with Reinforcement Learning.}
Motivated by this drawback, we design a reinforcement learning based framework to learn a policy for matching order generation.
We craft the long-term cumulative reward for the RL-based model to force the model to consider the overall computational cost when selecting the next node for matching order $\phi$.
Instead of only focusing on the current step, the RL-based framework allows the learned policy to consider the states multiple steps ahead before making the decision.
We also design an entropy reward to encourage the model to explore the unvisited states.
Instead of enumerating all possible orders and evaluating the quality of each of them, \mname\ samples a subset of such orders and learns the policy network from the observed rewards.
Accordingly, \mname\ has higher probability to produce matching orders that are closer to the optimal ones compared to the existing ordering methods with acceptable overhead.

\noindent \textbf{Contributions} 

The contributions of our work are summarized as follows:

\begin{itemize}
    
    \item To the best of our knowledge, our proposed \mname is the \textit{first} work to employ Reinforcement Learning technique to obtain 
    high-quality matching order for backtracking based subgraph matching algorithms.
    
    \item Our RL-based model could fully exploit the graph information and select the next query node 
    for the matching order at every step according to the cumulative reward, which can consider the long-term benefits.
    
    \item Extensive experiments conducted on $6$ real-life graphs demonstrate that the high-quality matching order
    obtained by \mname can significantly save the query processing time by up to two orders of magnitude. 
    
\end{itemize}

\noindent \textbf{Organization.}
The rest of the paper is organized as follows:
Section~\ref{sec:background} includes the preliminaries, definitions and related works.
Section~\ref{sec:model} presents the details of our proposed model \mname.
We report the experiment results in Section~\ref{sec:exp} and give a conclusion in Section~\ref{sec:conclusion}.
\section{Background and Related Work}
\label{sec:background}
In this Section, we introduce the preliminaries, problem statement, state-of-the-art subgraph matching algorithms and related works.

\subsection{Preliminaries}
\label{subsec:preliminary}

In this paper, we focus on the subgraph matching problem on the undirected labeled graph $G = (V, E)$, where $V$ denotes the vertices set, $E$ is a set of edges. There is a label function $f_l$ which maps a vertex $v \in V$ into a label $l$ in the label set $L$.
The query graph, denoted by $q$, is a potential subgraph of data graph $G$.
$q$ and $G$ share the same label function $f_l$ and the same label set $L$.
The frequently used notations are summarized in the Table~\ref{tb:notations}.

\begin{table}[tb]
\vspace{-2mm}
\caption{The summary of notations}
\small
  \centering
  \vspace{-2mm}
\scalebox{0.82}{

    \begin{tabular}{|c|l|}
      \hline
      \cellcolor{gray!25}\textbf{Notation} & \cellcolor{gray!25}\textbf{Definition}        \\ \hline
      $g$, $q$, $G$ &  graph, query graph and data graph.\\ \hline
      $V$, $E$, $L$ & vertex set, edge set and label set.\\ \hline
      $f_l$, $f_{iso}$ & label mapping function and subgraph isomorphism function.\\ \hline
      $e(u,v)$, $N(v)$ & an edge between node $u$, $v$ and the neighbor set of $v$. \\ \hline
      $d(u)$ & degree of node $u$. \\ \hline
      $C$, $LC$ & set of candidate vertices and local candidate vertices. \\ \hline
      $\phi$, $\#_{enum}$ & matching order and enumeration number. \\ \hline
      $N^\phi_+(u), N^\phi_-(u)$ & backward and forward neighbors of $u$ given $\phi$.\\ \hline
     \end{tabular}
}

\label{tb:notations}
\end{table}

We give the key formal definition of the subgraph isomorphism as follows:

\theoremstyle{definition}
\begin{definition}[Subgraph Isomorphism]
Given a query graph $q=(V,E)$ and a data graph $G=(V', E')$, a subgraph isomorphism is an injective function $f_{iso}$ from $V$ to $V'$ such that (1) $\forall v \in V, f_l(v) = f_l(f_{iso}(v))$; and (2) $\forall e_{(u,v)} \in E, e_{(f_{iso}(u), f_{iso}(v))} \in E'$.
\end{definition}

\noindent \textbf{Subgraph Matching.} The objective of the subgraph matching is searching for all subgraph isomorphisms from query graph $q$ to data graph $G$. Generally, the existing subgraph matching algorithms can be classified into two categories: backtracking based methods and join-based methods. Both directions have been intensively studied in the literature. In this paper, we focus on the backtracking based subgraph matching methods.

\begin{algorithm}[tb]
\small
\DontPrintSemicolon
  
  \KwInput{Query graph $q$, data graph $G$.}
  \KwOutput{Subgraph Match Mapping $M$ from $q$ to $G$.}
\tcp{The filtering method.}
$C \leftarrow$ generate candidate vertex sets\; 
\tcp{The ordering method.}
$\phi \leftarrow$ generate a matching order based on $q$, $G$ and $C$\;
\tcp{The enumeration procedure, finds the mapping $M$ for all query nodes in $q$.}
$M=$ \Enum{$q,G,C,\phi,\{\},1$}\;
\caption{Generic Subgraph Matching}
\label{al:generic}
\end{algorithm}

As summarized in previous
works~\cite{sun2020memory,DBLP:journals/pvldb/LeeHKL12}, backtracking based
subgraph matching method can be partitioned into three phases: (1) the
\textit{complete candidate vertex set} generation; (2) matching \textit{order}
generation; and (3) \textit{matching enumeration}. This general framework
(from~\cite{sun2020memory}) is outlined in Algorithm~\ref{al:generic}.

In more details, \textbf{Phase (1)} is to pre-process the graph where various filtering
techniques are deployed to generate candidate vertex sets $C$ for all query
nodes; this can reduce the search space before the enumeration process begins.

\begin{definition}[Complete Candidate Vertex Set $\mathcal{C}$]
  Given $q$ and $G$, a complete candidate vertex set $C(u)$ of $u \in V(q)$ is a
  set of data vertices such that for each $v \in V(G)$, if $(u, v)$ exists in a
  match from $q$ to $G$, then $v \in C(u)$.
\end{definition}

In \textbf{Phase (2)}, a match order $\phi$ is generated to guide the
enumeration of the matched subgraphs, which is formally defined as follows.

\begin{definition}[Matching Order]
A matching order $\phi$ is a permutation (i.e., sequence) of query graph's vertex set $V(q)$.
\end{definition}

\begin{definition}[Backward (Forward) Neighbors]
  With given order $\phi$, the backward (forward) neighbors $N^\phi_+(u)$
  ($N^\phi_-(u)$) are the neighbors of $u$ positioned before (after) $u$ in
  $\phi$.
\end{definition}

Usually, the matching order is generated heuristically based on the label
frequency, degree of vertices and other structural and attribute information.
For example, QSI introduces an \textit{infrequent-edge first ordering method},
VF2++ uses a \textit{infrequent-label first order} and CFL proposes a
\textit{path-based ordering method}.

Following is the formal definition of enumeration procedure in \textbf{Phase (3)}, which
finds all matches of the query subgraph $q$ in the data graph $G$.

\begin{definition}[Enumeration Procedure]
  An enumeration procedure is performed recursively to find subgraph
  matches $f_{iso}$ with given matching order $\phi$ and candidate vertex set
  $C$.
\end{definition}


\subsection{Problem Statement}
\label{subsec:pre_problem}


As highlighted in~\cite{sun2020memory}, it is critical to choose a good matching
order since the enumeration number in the search is significantly influenced by
the matching order, which is strongly correlated to the query time.
Following is a formal definition of the enumeration number.
\begin{definition}[Enumeration Number]
  An enumeration number $\#_{enum}$ is the number of \textit{recursive} calls of the 
  enumeration procedure to find all matches with given $q$, $G$,
  $\phi$ and $C$.
\end{definition}

It is cost-prohibitive to find an \textit{optimal} matching order, and existing
algorithms resort to heuristic strategies to find a good matching order.
Inspired by the great successes of RL techniques in a variety of optimization
problems~\cite{DBLP:conf/icml/BelloZVL17}, in this paper, we \textbf{aim to
  apply Reinforcement Learning technique to find a matching order} for given query graph $q$ and
data graph $G$ to \textbf{minimize the enumeration number} in the matching enumeration
phase such that the search performance of the subgraph matching can be enhanced.

\subsection{State-of-the-Art}
\label{subsec:SOTA}

In this subsection, we introduce the backtracking search based subgraph matching
algorithm \hname which is shown to achieve state-of-the-art search performance
according to the intensive empirical results~\cite{sun2020memory}, and the recent published algorithm VEQ~\cite{kim2021versatile}. 

\textbf{\hname} utilizes the candidate filtering, vertex ordering and enumeration methods of
GraphQL~\cite{DBLP:conf/sigmod/HeS08}, RI~\cite{bonnici2013subgraph} and
QuickSI~\cite{shang2008taming} respectively, which are introduced as follows:

\vspace{1mm}
\noindent \textbf{(1) Candidate Set Generation.} To limit the size of candidate vertex set for each node in the query, \hname
utilizes the candidate vertex filtering method in
GraphQL~\cite{DBLP:conf/sigmod/HeS08}.
Particularly, \hname generates the candidate set with local pruning and global refinement.
Local pruning filters out the invalid nodes based on the \textit{profile} of the neighborhood graph of the query vertex $u$, which is the lexicographic order labels of $u$ and neighbors of $u$.
If the profile of query vertex $u$ is a sub-sequence of that of data vertex $v$, then $v$ is added to $C(u)$.
Global refinement prunes $C(u)$ as follows:
given $v\in C(u)$, the algorithm first build a bipartite graph $B^v_u$ between $N(u)$ and $N(v)$ by adding $e(u', v')$ where $u'\in N(u)$ and $v'\in N(v)$, if $u'\in C(v')$.
Global refinement then checks whether there is a semi-perfect matching in $B^u_v$, \textit{i.e.,} whether all vertices in $N(v)$ are matched.
If not, $v$ will be removed from $C(u)$.

\vspace{1mm}
\noindent \textbf{(2) Matching Order Generation.}
Query vertex order generation is one of the key factors that reduce the query processing time.
\hname exploits RI's order generation method which is the state-of-the-art ordering method~\cite{sun2020memory}. 
\hname generates the
matching order only based on the structure of query graph $q$. \hname first
selects the query node with the maximum degree $u*= \argmax_{u\in V(q)} d(u)$ as
the starting nodes of order $\phi$. We denote the order determined at the $t$-th
step as $\phi_t$. Then the nodes with the most neighbors in $\phi_t$ are
iteratively selected, \textit{i.e.,}
$u* = \argmax_{u\in V(q) \wedge u \not\in \phi_t}\left|N(u)\cap \phi_t\right|$.
Since ties can easily occur in the $\argmax$ function, \hname breaks the ties by
considering following properties in order of: (1) the maximum value of
$|\{u'\in \phi_t \mid \exists u'' \in V(q) \setminus \phi_t, e(u',u'') \in E(q)
\land e(u, u'') \in E(q)\}|$, \textit{i.e.,} the number of vertices in $\phi_t$ that
have a neighbor outside of $\phi_t$ and is adjacent with $u$. The number is
denoted as $\left|u_{neig}\right|$. (2) the maximum number of neighbors of $u$
that are not in $\phi$, and not even adjacent with vertices in $\phi_t$, which is
defined as
$\left|u_{unv}\right|=\left|\{u'\in N(u) - \phi_t | \forall u'' \in \phi, e(u',u'')
  \not\in E(q)\}\right|$. If these values are still the same for at least two
nodes, \hname will choose a node arbitrarily.

Since \hname only considers the structure of query graph to build the matching order, it ignores the abundant information of the relationship between the query and data graphs. 
Consequently, \hname has to apply random selections in many cases, even selects the starting vertex arbitrarily.
This selection cannot guarantee to produce a robust matching order to reduce redundant intermediate results.


\begin{algorithm}[tb]
\small
\DontPrintSemicolon
  
  \KwInput{Candidate set $C$, query graph $q$, data graph $G$, match order $\phi$.}
  \KwOutput{Subgraph Match Mapping $M$.}
\Enum{$q,G,C,\phi,M,i$}\;
\tcp{Start with \Enum{$q,G,C,\phi,\{\},1$}}
\Begin{
    \If{$i>|\phi|$}{Output $M$, \textbf{return.}}
        $u$ $\leftarrow$ select an extendable vertex given $\phi$ and M\;
        \tcp{Compute local candidate set}
        $LC(u, M) \leftarrow ComputeLC(q, G,C,\phi,M,u , i)$\;
        \For{$v \in LC(u, M)$}
        {
        \If{$v \not\in M$}
        {
        Add $(u, v)$ to $M$\;
        $\mathrm{Enumerate}$($q,G,C,\phi,M,i+1$)\;
        Remove $(u, v)$ from $M$\;
        }
        }
}
\caption{Enumeration Procedure}
\label{al:enumeration}
\end{algorithm}

\vspace{1mm}
\noindent \textbf{(3) Enumeration Procedure.}
\hname adopts the recursive enumeration procedure which is also widely used in existing backtracking search based methods.
The enumeration procedure is summarized in Algorithm~\ref{al:enumeration}.
The function \Enum{$\cdot$} recursively enumerates all results.
$M$ is a set that records all mappings from query vertices to the data graph vertices.
Once all the vertices of the query graph are mapped, Line $4$ outputs the mapping $M$.
Otherwise, the algorithm selects an extendable vertex which is defined as follows:

\begin{definition}[Extendable Vertices]
Given matching order $\phi$ and mapping $M$, extendable vertices $\Gamma(\phi, M)$ are query vertices $v$ such that each $u'\in N^{\phi}_+(u)$ has been mapped in $M$ but $u$ has not.
\end{definition}

  

Line 6 shows the computation of the local candidate set after selecting the extendable vertex.
The local candidate set is computed following different rules for specific subgraph matching algorithms.
\hname generates the local candidate set $LC(\cdot)$ by checking whether the vertices in candidate set are connected to the last matched data vertices.
Line 7-11 loop over $LC(u, M)$ to find more mappings to extend $M$, and recursively invoke the \Enum{$\cdot$} function.
Finally, we get the matching from $q(\phi(1:i))$ to $G$.

\textbf{VEQ}~\cite{kim2021versatile} is a recent published algorithm for subgraph matching problem.
VEQ follows the three-phase framework mentioned above.

\vspace{1mm}
\noindent \textbf{(1) Candidate Set Generation.}
VEQ first builds the query directed acyclic graph (DAG) of query graph $q$ by assigning directions to edges in $q$.
VEQ also finds the neighbor equivalence class (NEC) for degree-one vertices, in which the query vertices have the same label and the same neighbors. 
VEQ then builds the candidate set by using extended DAG-graph dynamic programming with an additional filtering function that utilizes a concept called neighbor-safty.

\vspace{1mm}
\noindent \textbf{(2) Matching Order Generation.}
VEQ generates the matching order based on the size of candidate set and the size of neighbor equivalence class (NEC), which could reduce the redundancy in search space.
However, this order generation method is still based on greedy heuristics and might be trapped in local optimum.

\vspace{1mm}
\noindent \textbf{(3) Enumeration Procedure.}
VEQ follows the same enumeration procedure introduced above.
Besides, VEQ prunes out repetitive subtrees of the search space by utilizing dynamic equivalence of the subtrees.



In this paper, we design the RL-based matching order generation method to improve the overall search performance.
Specifically, we utilize the same candidate set generation and enumeration methods as \hname.


\begin{figure*}[tbh]
    \centering
    \includegraphics[width=0.85\textwidth]{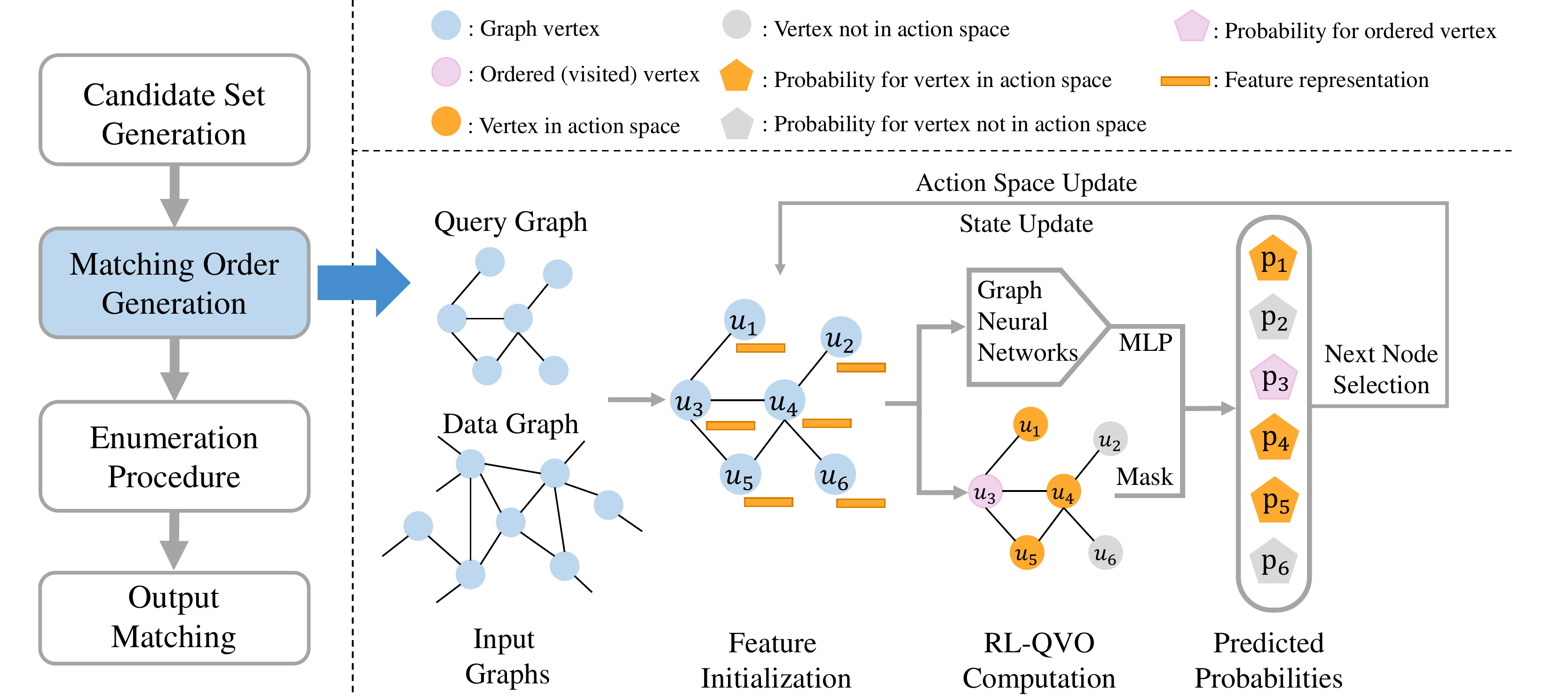}
    \vspace{-1mm}
    \caption{Framework of \mname}
    \vspace{-4mm}
    \label{fig:framework}
\end{figure*}

\subsection{Related Work}
\label{subsec:related}
In this subsection, we introduce the works which are closely related to your study.

\vspace{1mm}
\noindent \textbf{Subgraph Matching.}
There are two major categories of subgraph matching methods.
There are great number of subgraph matching models can be categorized as backtracking search based algorithms~\cite{shang2008taming,DBLP:conf/sigmod/BiCLQZ16,han2013turboiso,DBLP:conf/sigmod/HanKGPH19,DBLP:journals/pami/CordellaFSV04,DBLP:conf/gbrpr/CarlettiFV15,juttner2018vf2++,carletti2017challenging,kim2021versatile}.
These models follow the filtering, ordering and enumeration framework to obtain matches for query graphs.
The other category is the join-based methods~\cite{DBLP:journals/pvldb/LaiQLC15,DBLP:journals/pvldb/LaiQLZC16,yang2021huge} which usually aim at the subgraph enumeration problem.
These algorithms convert subgraph enumeration problem to a multi-way join task, which are also utilized in many database systems~\cite{DBLP:conf/pods/000118,DBLP:conf/sigmod/ArefCGKOPVW15}.
In this paper, we aim at designing an ordering method for backtracking search based algorithms.
Besides, there is a family of subgraph matching methods~\cite{zampelli2010solving,mccreesh2020glasgow,mccreesh2018subgraph} which are based on constrained programming.
These methods are proven proven optimal for several mid-sized or highly complicated cases.
There are also several algorithms~\cite{zhang2009gaddi,zhao2010graph,rivero2017efficient} utilize indexing-enumeration framework, which constructs indices on $G$ and answers all queries with the assistance of the indices.
These index-based methods would outperform the non-index-based methods in many cases, please refer to~\cite{katsarou2015performance} for detailed experimental survey.
However, these algorithms may have severe scalability issues according to~\cite{lee2012depth,sun2020memory,katsarou2015performance}.

Recently, there have been several research attempts~\cite{lou2020neural,liu2020neural,chen2020can} to develop subgraph matching algorithms on a learning basis.
These methods focus on counting of isomorphic subgraphs or producing approximate matching results, which are different from the target in this paper, \textit{i.e.,} generating matching orders to find exact subgraph mappings.



\vspace{1mm}
\noindent \textbf{Reinforcement Learning.}
Using reinforcement learning for graph-related problems is becoming increasingly popular.
\cite{dai2017learning,kurin2019improving,li2018combinatorial,DBLP:conf/nips/GasseCFC019,DBLP:journals/www/RenAGWG20} aim to solve the NP-hard combinatorial optimization problems with reinforcement learning and graph representation learning techniques.
Please see~\cite{bengio2020machine} for detailed survey.
Several RL-based models have been proposed for graph analysis tasks.
For example, \cite{DBLP:conf/nips/HuXQYC0T20} exploits a GNN-based policy network in promoting the performance for cross-domain graph analysis tasks such as classification; \cite{DBLP:conf/nips/YouLYPL18,DBLP:conf/kdd/Do0V19} utilize graph policy networks on the molecular graphs for chemical and medical applications such as chemical reaction prediction. 

\vspace{1mm}
\noindent \textbf{Graph Neural Networks.}
Graph neural networks (GNNs) have shown great success in various graph-related applications~\cite{wang_gognn_2020,hao2020inductive,hao2021ks,wang2021binarized,wang2021powerful,10.1145/3494558}.
Typically, GNNs learn the node representations with iterative aggregation of neighborhood information for each node, e.g., GCN~\cite{DBLP:conf/iclr/KipfW17}, GAT~\cite{DBLP:conf/iclr/VelickovicCCRLB18} and GraphSAGE~\cite{hamilton2017inductive}.
Despite these models, there have been a great number of GNNs proposed for wide range of real-life applications, please refer to~\cite{wu2020comprehensive} for comprehensive survey.


\section{Our Approach}
\label{sec:model}

In this section, we present a reinforcement learning based model \mname to generate high-quality matching orders.

\subsection{Motivation}
\label{subsec:alg_motivation}

The backtracking-based subgraph matching methods explore the search space with a given matching order, 
which presents several unique challenges that are good fits for reinforcement learning and graph neural network:

First, it is cost-prohibitive to find the optimal matching order for a given query graph $q$ and a data graph $G$.
The existing ordering methods adopt the greedy heuristics to pick nodes that could possibly reduce the search space 
at current step as early as possible. 
However, these selections may result in the global \textit{suboptimality}.
With the help of reinforcement learning technique, our model is able to generate the matching order by considering the long-term rewards multi-steps ahead.
Therefore, our proposed \mname\ has better chance to escape from local optimum, and eventually shows higher matching order quality than the greedy methods.

Second, existing ordering methods only utilize the local neighborhood information with fixed priority to produce the matching order, which has a non-negligible probability to ignore some structural and attribute information within $q$ and $G$.
Graph neural networks (GNNs) have already been proven to have powerful ability of exacting information for specific graph analysis tasks. 
With the reinforcement learning framework, \mname\ does not need to make assumptions on the query or data graph distributions, but regards the ordering process as a black-box computation on representations learned by GNNs with carefully designed initial features.
As a result, \mname\ generates the matching order adaptively with regards to different query graphs.



These motivate us to apply the Graph Neural Networks and Reinforcement Learning techniques to better utilize the graph information and find high-quality matching orders by looking multi-steps ahead (i.e., long-term rewards) during the training process. 
Though the training of the model needs extra overhead compared to existing matching order generation approaches, it can be preprocessed which is a common practice for various indexing techniques in the literature.
Moreover, we show that the generation of the matching order during the query processing is very time efficient,
and the gain is significant as demonstrated in our experiments. 

There are three major categories of reinforcement learning algorithms: value function based methods, policy search methods and the methods combine both value function and policy search such as actor-critics~\cite{arulkumaran2017deep}.

We observed in our initial experiments that the enumeration numbers for the query vary vastly with different matching orders.
Therefore, the methods use value function, such as Q-learning and actor-critics, are hard to converge, which leads to unsatisfactory results.
According to these observations, we choose to apply policy search method in our model.

\subsection{Framework}
\label{subsec:alg_framework}

The research focus of this paper is to design a novel model to generate good matching orders
for backtracking based subgraph matching algorithms.
Thus, we replace the matching order generation part of the the state-of-the-art algorithm \hname
with a new model: \textbf{Reinforcement Learning Based Query Vertex Ordering Model} (\textbf{\mname} for short).

Figure~\ref{fig:framework} illustrates the framework of the \mname.  
Particularly, \mname\ regards the matching order generation as a Markov Decision Process (MDP),
and the vertices in a matching order are generated sequentially.
At the beginning, \mname\ first aggregates neighbor information with carefully designed features to produce node representations for each query vertex. 
With the obtained representations, \mname\ computes the probability scores for query vertices to
guide the selection of the next query node for the matching order.
The matching order is output if all query nodes are processed. Otherwise, we will repeat this procedure with
updated action space and features. 
More details of the model will be introduced in the following subsections.

\subsection{Query Vertex Ordering as Markov Decision Process}
\label{subsec:alg_MDP}

The query vertex ordering process can be naturally formulated as a Markov Decision Process (MDP).
Given ordered nodes $\phi_t$ and input feature matrix $\bm H^t$ of query graph at step $t$ as a state,
reinforcement learning method takes an action from action space determined by the neighbors of the ordered vertices $N(\phi_t)=\{N(u)|\forall u \in \phi_t, N(u) \not\in \phi_t\}$ with predicted probabilities to select next node for the order $\phi_{t+1}$.
After every selection, the feature matrix $\bm H$ and action space are also updated.
With the generated order $\phi$, the model performs the enumeration procedure and gets the reward based on the reduced number of enumerations $\Delta \#_{enum}$ compared to baselines, entropy of predicted probabilities and number of valid predictions.
The MDP is formally defined as follows:

\noindent \textbf{State.}
At step $t$, the state is defined by the order $\phi_{t}$ which contains $t$ vertices and query graph representation matrix $\bm H_q^t$.
The whole query graph feature matrix $\bm H_q$ is involve in the state representation in order to familiarize the model with the overall structural and attributed information of the query graph.
Since \mname\ is a machine learning-based model, for each vertex in query graph, we carefully initialize a feature representation ${\bm h}_u^{(0)}$.

\noindent \textbf{Feature Representations.}
As discussed in Section~\ref{subsec:preliminary}, the statistical heuristics of query graphs, such as degree of query vertices, label frequency and \textit{etc}, play a key role in determining the query vertex order.
Inspired by~\cite{DBLP:conf/nips/HuXQYC0T20}, we initialize the input features for query vertices based on important heuristics in order to fully exploit the information and relations hidden in and between query and data graphs.
Specifically, we generate the initial feature as follows:

We compute the degree of each node. Intuitively, the query node with greater degree will have less matching in the data graph, and eventually has higher priority to be added in the order $\phi$.
Therefore, we use a scaled degree as one of the initial state value, \textit{i.e.,}
\begin{equation*}
    {\bm h}_u^{(0)}(1) = degree(u)/\alpha_{degree},
\end{equation*}
where $\alpha_{degree}$ is a scaling factor to ensure the computation stability.

The initial feature also includes the node label information.
Since integers are used to represent labels in our data format, we simply put the integer label id into the representation for each query vertex.
\begin{equation*}
    {\bm h}_u^{(0)}(2) = label(u)
\end{equation*}

To enable the graph neural network used in \mname\ to discriminate the order of input node, we directly put the query node id in the initial representation.
\begin{equation*}
    {\bm h}_u^{(0)}(3) = id(u)
\end{equation*}
Please note that the query graph usually has small number of nodes, therefore there is no need to perform scaling on the vertex id.

We further include the data graph related heuristics to build our initial feature representation.
The following statistics are commonly used in existing models: the frequency of vertices $v$ in the data graph $G$ with greater degree than query vertex $u$; and the frequency of vertices $v$ in the data graph that have the same label as query vertex $u$.
They are vectorized in the initial feature representation.
\begin{equation*}
    {\bm h}_u^{(0)}(4) = \left|\{v\in G|d(u)<d(v)\}\right|/(\left|V(G)\right|\times\alpha_{d});
\end{equation*}
\begin{equation*}
    {\bm h}_u^{(0)}(5) = \left|\{v\in G|L(u)=L(v)\}\right|/(\left|V(G)\right|\times\alpha_{l});
\end{equation*}
where $\alpha_{d}$ and $\alpha_{l}$ are the scaling factors.
Please note that the frequency computations of ${\bm h}_u^{(0)}(4)$ and ${\bm h}_u^{(0)}(5)$ involve all data vertices, which reflect the distribution of the data graph.

Lastly, we put the number of unordered vertices and a trailing indicator at the initial representation.
The number of unordered vertices enables the model to make different decision at different time step $t$, which is formulated as:
\begin{equation*}
    {\bm h}_u^{t}(6) = \left|V(q)\right| - t + 1
\end{equation*}
The indicator variable is $0$ if the vertex $u$ has not been ordered before the selection at time step $t$ (\textit{i.e.,} $u$ is not in the order $\phi_{t-1}$) and $1$ otherwise, \textit{i.e.,}
\begin{equation*}
    {\bm h}_u^{t}(7) = \mathbbm{1}(u \in \phi_{t-1})
\end{equation*}
Before the selection starts, \textit{i.e.,} when $t=0$, indicators are $0$ for all query vertices, and the indicator variable will be updated with the change of the state at every time step.

We concatenate all these features to formulate the input representation.
This representation could easily incorporate additional heuristic features by appending these features to our initial feature representation vector.
In our model, we only use five primary heuristics introduced above with two trailing indicators to capture the differences between the query vertices, and exploit the reinforcement learning and graph neural networks to automatically learn more complicated and informative criterion for query vertex order generation.

\noindent \textbf{Action.}
The action space is defined by the neighbor vertices set $N(\phi_t)=\{N(u)|\forall u \in \phi_t, N(u) \not\in \phi_t\}$ of the ordered vertices at current step $t$.
At every step $t$, the action is to select vertex $u'$ from $N(\phi_t)$ according to the predicted probabilities and add $u'$ into the matching order for next step, \textit{i.e.,} $\phi_{t+1} = \phi_t\cup \{u'\}$.
Instead of directly selecting the vertex with greatest probability, \mname\ makes the selection according to the probabilities of vertices in the action space to allow more exploration.
Please note that the first vertex can also be selected according to the degree.

\noindent \textbf{Reward Design.}
The reward is key factor in reinforcement learning, in this paper, the reward includes the reduced number of enumeration,  validate reward for probability scores and entropy of the probabilities.
One immediate reward is the reduced enumeration number $\Delta \#_{enum} = \#_{enum}(\phi) - \#_{enum}(\phi_{base})$, where $\phi$ is the learned order of RL-based agent and $\phi_{base}$ is the baseline order produced by existing subgraph matching algorithms.
In our model, we use the state-of-the-art method \hname as our baseline algorithm.
Specifically, according to~\cite{sun2020memory}, the ordering method of RI~\cite{bonnici2013subgraph} has the best performance and is used in \hname.
Therefore, we choose the order produced by RI as our baseline order, \textit{i.e.,} $\phi_{base} = \phi_{RI}$.
Considering the varying orders of magnitude of $\Delta \#_{enum}$ with different query graphs, the enumeration reward $r_{enum}$ is defined as $r_{enum} = f_{enum}(\Delta \#_{enum})$, where $f_{enum}(\cdot)$ is a function such as logarithm which reduces the gaps (differences) between enumeration rewards under different query graphs to stabilize the computation.
Intuitively, the model is more likely to gain greater enumeration reward $r_{enum}$ on the complex queries which inherently require more rounds of enumeration procedure.
Actually, the average enumeration time is usually dominated by the time costs of these complex queries.
Therefore, the policy network pays greater importance to the complex queries with the hope to reduce more enumeration numbers in total.
Since the enumeration reward $r_{enum,t}$ at step $t$ cannot be determined until the final order $\phi$ is obtained, all rewards $r_{enum,t}$ at steps $t$ share the same value as $r_{enum} = f_{enum}(\Delta \#_{enum})$.
Meanwhile, this shared reward value enables the policy network to consider the long-term reward at every step, thus reduces the probability to fall into the local optimum.

We also design the step-wise validate rewards $r_{val,t}$.
A small positive reward is assigned if the policy network produces a valid probability distribution, i.e., the vertex with largest probability is in the action space $u' \in N(\phi_t)$.
Otherwise, a negative punishment is assigned which is greater than the positive reward in absolute value.
Please note that even if the policy network produces invalid probabilities, our model still guarantees to generate a \textit{connected order} $\phi$ by masking out the vertices that are not in the action space before making selection.

Furthermore, an entropy reward is considered at every step to encourage the model to output action probability distribution with \textit{high} entropy~\cite{ziebart2008maximum}.
Hence, the model is more likely to take actions unpredictably, which avoids the network converging too quickly on a policy that is locally optimal. 
This entropy reward $r_{h,t}$ is defined as $r_{h,t} = H(P_{\pi_{\theta}}(\phi_t, N(\phi_t))$, where $H(\cdot)$ is the entropy function, $\pi_{\theta}$ is a policy network with parameters $\theta$, and $P_{\pi_{\theta}}(\phi_t, N(\phi_t))$ is the output probability at step $t$ with given order $\phi_t$ and action space $N(\phi_t)$.
Therefore, overall step-wise reward is formulated in the following Equation~\ref{eq:reward}:

\begin{equation}
\label{eq:reward}
    R_t = r_{enum} + \beta_{val}\cdot r_{val,t} + \beta_{h}\cdot r_{h,t},
\end{equation}
in which $\beta_{val}$ and $\beta_{h}$ denote the reward coefficients for validate reward and entropy reward respectively to tune their impact on training process.

In query ordering for subgraph matching task, the starting nodes in the order are usually more important than the trailing nodes.
Therefore, when calculating the overall rewards for the policy network, we assign a decay factor to the step-wise rewards and formulate the overall rewards as follows:
\begin{equation}
    R_{q,\theta} = \sum_{t=1}^{\left|V(q)\right|}\gamma^tR_t,
    \label{eq:all_reward}
\end{equation}
where $\gamma \in (0, 1)$ is a decay factor which enables the model to consider the importance of nodes in the learned order during training procedure.

\subsection{RL-QVO Policy Network Architecture}
\label{subsec:alg_networks}

In this section, we introduce the architecture of policy network in our RL-based method \mname, which generates the query vertex order for subgraph matching.

\noindent \textbf{Framework.} Framework of \mname\ is illustrated in Figure~\ref{fig:framework}.
\mname\ first computes the vector representations $\bm{x}_u$ for query vertices $u \in V(q)$ by exploiting graph neural networks.
The representation matrix of query vertices $\bm{X}_q$ is served as the input of a multi-layer perceptron (MLP) to obtain the probability distribution on action space for vertex selection.

\noindent \textbf{Action Space.}
As introduced before, the action space $AS(t)$ is defined by the neighbor vertices set $N(\phi_t)=\{N(u)|\forall u \in \phi_t, N(u) \not\in \phi_t\}$ of the ordered vertices at current step $t$ to ensure the connectivity of generated orders.
This constraint applies for most ordering methods in backtracking based subgraph matching algorithms.
In the case that there is only one vertex in the action space, \textit{i.e.,} $\left|AS(t)\right|=1$, \mname\ directly selects the only candidate as the next node without performing computation.

\noindent \textbf{Policy Network.}
The policy network is a neural network which learns the rule for solving the target problem.
With given state, a policy network returns a probability distribution over the action space.
Then, according to the produced probability distribution, \mname\ progressively selects the vertex to add into the matching order $\phi$.

One naive solution is to use the simple multi-layer perceptron (MLP) as the policy network.
However, MLP cannot capture the structural information with in the query graphs and usually has limited performance in complex tasks.

In our model, the policy network includes two main parts: the graph neural network that aggregates and extracts the graph information; and the multi-layer perceptron which finally produces the probability distribution.

In order to obtain the vector representations for query vertices, we utilize graph neural network, a well-studied technique which achieves the state-of-the-art performance in graph representation learning.
We find in our experiment (see Section~\ref{subsec:ablation}) that the performance of our model is not bound to the selection of GNN.
Specifically, the policy network $\pi$ is parameterized as graph convolutional network (GCN)~\cite{DBLP:conf/iclr/KipfW17} to embed the query vertices.

The high level idea of graph convolutional neural network is to perform a message passing along edges in the graph for total of $\mathcal{L}$ layers.
Therefore, \mname\ could not only obtain the heuristic information from the initial feature matrices, but also make prediction based on auxiliary structural and high-order neighboring information which are overlooked in conventional subgraph matching algorithms.
The aggregation of graph convolutional neural network can be formulated as follows:

\begin{equation}
    {\bm H}^{(l+1)} = \sigma(\Tilde{\bm D}^{-\frac{1}{2}}\Tilde{\bm A}\Tilde{\bm D}^{-\frac{1}{2}}{\bm H}^{(l)}{\bm W}^{(l)}),
\end{equation}
where ${\bm W}^{(l)}$ and ${\bm H}^{(l)}$ are the weight and the input feature matrices of $l^{th}$ layer respectively, ${\bm H}^{(0)}$ is the initial feature representation introduced in Section~\ref{subsec:alg_MDP}. 
$\Tilde{\bm A} = {\bm A} +{\bm I}$ is the adjacency matrix with self loops, $\Tilde{\bm D}$ is a diagonal matrix where $\Tilde{\bm D}_{ii} =  \sum_i \Tilde{\bm A}_{ij}$, and $\sigma$ is an activation function such as ReLU.
\mname\ is compatible to any graph neural network.
Because GCN could capture the structural, label and neighboring information with simple network structure and efficient computation process, we choose graph convolutional network as the graph representation learning module of \mname.
$\mathcal{L}$ layers of GCNs output the representations $\bm H^{\mathcal{L}}_q$ for all query vertices.
\mname\ then applies the multi-layer perceptron (MLP) on obtained representation ${\bm H}^{\mathcal{L}}_q$ to select the next node.
Specifically, we use two linear neural layers with mask and normalization operations:
\begin{equation}
    \mathbb{P}_{u'}^{(t)} =\pi(\cdot|S^{(t)})= Softmax(mask_{u'\in AS(t)}({\bm W}_2 \cdot \sigma({\bm W}_1 {\bm h}^{(t)}_{u'}))),
    \label{eq:prob}
\end{equation}
Where ${\bm h}^{(t)}_{u'}$ is the vector embedding of query vertex $u'$.
The output dimension of the MLP is 1, therefore, we get a real number score as the selection probability for each node.
We utilize the mask operation to filter out the probability scores of vertices that are out of the action space, and then apply the softmax function on the candidate query vertices: $Softmax(\bm z) = \frac{e^{{\bm z}_i}}{\sum_j(e^{{\bm z}_j})}$ to produce the normalized probabilities.
These normalized probabilities for candidate vertices are also used to compute the entropy
rewards mentioned in Section~\ref{subsec:alg_MDP}.
Please note that in Equation~\ref{eq:prob}, we omit the bias for simplicity.


\subsection{Policy Training}
\label{subsec:alg_policy}

In the training phase, given $N_{T}$ training query graphs $q = \{q_i|i = 1, ..., N_T\}$ and a data graph $G$, our goal is to maximize the expected rewards of policy network $\pi_\theta$ for the training graphs.
As introduced in the previous subsection, the policy network is consist of layers of GCN and MLP.
The reward of \mname's policy network $\pi_\theta$ with parameters $\theta$ at time step $t$ is the summation of rewards for all query graphs in the training batch as follows:
\begin{equation}
    r_t(\theta) = \sum^{N_{T}}_{i=1}R_i(\phi_i;\theta_i),
\end{equation}
where $R_i$ is the reward for query graph $q_i$ defined in Eq.~\ref{eq:all_reward}.
We utilize the proximal policy optimization (PPO)~\cite{schulman2017proximal} to train the policy network.
PPO is a policy gradient method which utilizes a sampling policy network that enables the model to update with sampled data in multiple epochs.
In our case, the policy network $\pi_{\theta'}$ from the previous epoch (has not been updated) is used as the sampling policy network.
The loss function is formulated as
\begin{align}
\label{eq:loss_0}
    \medmath{J^{(t)}_r(\theta) = \sum_{(a_t,s_t)} min(\frac{\pi_\theta(a_t|s_t)}{\pi_{\theta'}(a_t|s_t)}r_t(\theta), clip(\frac{\pi_\theta(a_t|s_t)}{\pi_{\theta'}(a_t|s_t)}, 1-\epsilon, 1+\epsilon)r_t(\theta)),}
\end{align}
\begin{equation}
\label{eq:loss}
    J(\theta) = \sum_{t=1}^{\left|V(q)\right|}J^{(t)}_r(\theta),
\end{equation}
where $\theta'$ is the parameters of policy network in previous epoch, $\epsilon$ is a factor to clip $\frac{\pi_\theta(a_t|s_t)}{\pi_{\theta'}(a_t|s_t)}$ which is the probability ratio of action $a_t$ with state $s_t$ computed by current and sampling policy networks.
At every training step, the parameters from the previous epoch $\theta'$ remain fixed, and the parameters $\theta$ of the policy network are updated via backpropagation according to loss function described in Eq. (\ref{eq:loss_0}), (\ref{eq:loss}).
The model replaces the parameters of sampling policy network with that of current policy network after each training epoch.

With the matching order obtained by the policy network, the algorithm executes the enumeration procedure to find all subgraph matches.
We adopt the commonly used enumeration procedure summarized in Algorithm~\ref{al:enumeration}, which is also utilized in the state-of-the-art baseline method \hname.


\subsection{Incremental Training}
\label{sec:incr_train}
In experiment, we found that \mname might have poor performance on small query sets, because the query sets with more vertices since these graphs usually require more enumeration number and potentially have greater values for objective function, which dominate the time cost of training procedure.
Besides, more training time is required if we train \mname on all query sets.
Therefore, we incorporate the incremental training procedure which first trains \mname on one query set for a large number of epochs and then trains \mname on other query set by minimizing objective function in Eq. (\ref{eq:loss}).
In our experiment, we found a smaller value of training epochs (e.g., 10) is enough to prevent \mname from catastrophic forgetting, improve the performance of \mname on new query set and save the overall training time.
Please refer to Section~\ref{subsec:incre_exp} for more details.

\subsection{Complexity Analysis}
\label{subsec:alg_complexity}

\mname\ is built on the neural network framework that enjoys excellent time efficiency in terms of query process.
With a given query graph $q$ and data graph $G$, \mname\ is only required to perform computation of graph neural networks and MLP on the query graph to produce the probability for node selection at every time step.
In order to obtain the matching order for graph with $\left|V(q)\right|$ vertices, such computation should be executed for $\left|V(q)\right|$ times.
Since the time complexity of GNN is $O(\lvert E(q) \rvert)$~\cite{wu2020comprehensive} and that for MLP is $O(d^2)$ where $d$ is the dimension of representations.
As a result, the overall time complexity of \mname\ for query vertex ordering is $O(\lvert V(q) \rvert \times (\lvert E(q) \rvert + d^2))$.
Because the size of query graph $q$ is relatively small, the time complexity of \mname\ is negligible compared to the time cost of recursive enumeration.
In terms of the space complexity, \mname\ requires fixed space for the parameters, which is determined by the input and output vector dimensions of a network.
Specifically, the space complexity is $O(\mathcal{L}\times d^2)$, where $\mathcal{L}$ is the number of neural layers and $d$ is the dimension of representations.
Therefore, our proposed \mname\ remains fixed space requirement with growing sizes of query and data graphs.

\subsection{Discussion}
\label{subsec:discussion}
In the training phase of this work, we adopt the Proximal Policy Optimization (PPO) to train our policy network.
Though PPO outperforms other reinforcement learning training methods, such as actor-critic and Q-learning in this work, it also requires the matching on the training instances, which results in extra training time.
The training time is significantly reduced by incremental training and reducing number of enumerated matches in the training phase.
This overhead could be entirely avoided in other reinforcement learning frameworks.
This could be an opportunity for the future work.

\section{Experiment}
\label{sec:exp}
This section shows the results of empirical studies.
\subsection{Experiment Setup}
\label{sec:setup}
\noindent \textbf{Compared Methods.}
In order to show the efficiency and effectiveness of our proposed \mname, we conduct the experiments on the following compared methods including \hname whose techniques are recommended in~\cite{sun2020memory}.
\begin{itemize}
    \item \textbf{QuickSI (QSI)}~\cite{shang2008taming} is a directly enumeration method which does not generate candidate set, but filters out the unpromising vertices during enumeration. QSI uses the \textit{infrequent-edge first ordering method}.
    \item \textbf{RI}~\cite{bonnici2013subgraph} is a state space representation based model which generate the query vertex order only based on the structure of query graph $q$.
    \item \textbf{VF2++}~\cite{juttner2018vf2++} is also a state space representation based model which generates the query vertex order according to the node label frequency.
    \item \textbf{VEQ}~\cite{kim2021versatile} is a recent published method with state-of-the-art performance on both subgraph query and subgraph matching tasks.
    \item \textbf{Hybrid}~\cite{sun2020memory} has a combination of candidate filtering, vertex ordering and enumeration methods of GQL, RI and QSI~\cite{veldhuizen2012leapfrog} respectively.
    These techniques are proven to have the best performance according to an extensive empirical study~\cite{sun2020memory}.
    \item \textbf{\mname} is our proposed backtracking based subgraph matching algorithm. Particularly, the proposed RL-based method is utilized to generate the matching order.
    Regarding the candidate set generation and enumeration procedure, we use the same implementations of \hname.
\end{itemize}

We obtained the code of VEQ~\footnote{\url{https://github.com/SNUCSE-CTA/VEQ}} from the authors of~\cite{kim2021versatile}.
All other baseline methods are implemented by the authors of~\cite{sun2020memory} in C++~\footnote{\url{https://github.com/RapidsAtHKUST/SubgraphMatching}}.
The machine learning part of \mname\ is implemented using Pytorch which could be computed on GPUs, while the candidate generation and enumeration methods are adapted from the code of~\cite{sun2020memory} in C++.

\noindent \textbf{Experiment Environment.}
We conducted the experiments on the servers running RHEL 7.7 system, which have Intel Xeon Gold 6238R 2.2GHz 28cores CPU and NVIDIA Quadro RTX 5000 GPU with 88GB RAM (Six Channel).

\noindent \textbf{Data Graph.} 
We conducted the experiments on six real-life datasets.
The number of vertices varies from $3,112$ to $1,134,890$.

\begin{table}[tb]
\centering
\caption{Datasets Properties}
\label{tb:data}
\begin{tabular}{|c|c|c|c|c|}
\hline
\textbf{Dataset}   & \textbf{$|V|$} & \textbf{$|E|$} & \textbf{$|L|$} & \textbf{d} \\ \hline
\textbf{Citeseer}    & 3,327        & 4,732        & 6            & 1.4        \\ \hline
\textbf{Yeast}     & 3,112        & 12,519       & 71           & 8.0        \\ \hline
\textbf{DBLP}      & 317,080      & 1,049,866    & 15           & 6.6        \\ \hline
\textbf{Youtube}    & 1,134,890    & 2,987,624    & 25           & 5.3        \\ \hline
\textbf{Wordnet}   & 76,853       & 120,399      & 5            & 3.1        \\ \hline
\textbf{EU2005}     & 862,664      & 16,138,468   & 40           & 37.4       \\ \hline
\end{tabular}
\vspace{-3mm}
\end{table}

We obtain the datasets used by previous works~\cite{sun2020memory,DBLP:conf/sigmod/BiCLQZ16}.
Six real-life graphs can be classified into five different categories: Citeseer is a citation network, Yeast is a biology network, DBLP and Youtube are social networks, Wordnet is a lexical network and EU2005 is a web network.
The detailed properties of the datasets are summarized in Table~\ref{tb:data}.


\noindent \textbf{Query Graph.}
As described in \cite{sun2020memory}, the query graphs are generated for each data graph by randomly extracting \textbf{connected} subgraphs from G.
We generate the query set for Citeseer following the setting introduced in~\cite{sun2020memory}.
The number of query vertices $\left|V(q)\right|$ varies from 4 to 32 for Citeseer, Yeast DBLP, Youtube and EU2005; $\left|V(q)\right|$ varies from 4 to 16 for Wordnet.
There are 400 query graphs in $Q_8$ and $Q_{16}$, and 200 query graphs in $Q_4$ and $Q_{32}$.
$50\%$ of the query graphs are used for training, and the remaining query graphs are used for evaluation.
Table~\ref{tb:query} lists the details of query sets for each dataset, where $Q_i$ denotes the query set in which the query graphs have $i$ vertices.
Due to the space limit, by default, we report the time cost results on query graph set with $16$ vertices, \textit{i.e.,} $Q_{16}$ for Wordnet, and query set with $32$ vertices, \textit{i.e.,} $Q_{32}$ for other data graphs as representatives.


\begin{table}[tb]
\centering
\caption{Query Sets}
\vspace{-2mm}
\begin{tabular}{|c|c|c|}
\hline
\rowcolor[HTML]{C0C0C0} 
Dataset                           & Query Set            & Default \\ \hline
\begin{tabular}[c]{@{}l@{}}Citeseer, Yeast, DBLP, \\Youtube, EU2005 \end{tabular} & $Q_4$, $Q_8$, $Q_{16}$, $Q_{32}$ & $Q_{32}$
\\\hline
Wordnet & $Q_4$, $Q_8$, $Q_{16}$ & $Q_{16}$  \\ \hline
\end{tabular}
\label{tb:query}
\vspace{-5mm}
\end{table}


\noindent \textbf{Experiment settings.}
In this experiment, \mname\ has two layers of graph convolutional networks to obtain the representations for the query nodes.
After GCN layers, there is a two-layer linear neural network that produces the selection probability for each node.
The learning rate is set to $10^{-3}$, the output dimension of GCN is set to 64, number of training epochs is 100 by default and 10 for incremental training.
All the scaling factors $\alpha$ are set to 1.
We also apply a dropout ratio at 0.2 during training.
Due to the enormous time cost to find all subgraph matches, existing works~\cite{sun2020memory,kim2021versatile} measure the matching time to find the first $10^5$ matches.
In this work, unless otherwise stated, we follow the settings of the existing works and terminate the enumeration procedure when $10^5$ matches are found during training and evaluation phases for fair comparison and reducing overall time cost.
We set a time limit $500$ seconds for subgraph matching, if the query process exceeds the time limit during training, we skip this query graph for training to save experimental time.
During evaluation, if a compared algorithm cannot finish the query within the time limit, we denote the query graph as a unsolved query and assign the time cost as $500$ secs for this query.
If a query graph remains unsolved by all compared methods, we would exclude this query graph in computing the average 
query processing time and enumeration time.
We also count the numbers of unsolved queries to compare the capabilities in solving the worse case queries for all compared methods.
The results are shown in Section~\ref{sec:time_cost}.

\noindent \textbf{Evaluation Metrics.}
In this experiment, we evaluate the time efficiency of compared methods.
We use the average query processing time (i.e., the query response time) and enumeration time (i.e., the time cost of enumeration procedure) for the comparison.
Recall that we say a query is unsolved if it cannot be finished within $500$ seconds. 
We also report the training time and the memory consumption of the our model.


\subsection{Query Processing Time Comparison}
\label{sec:time_cost}

\begin{figure}[tb]
    \centering
    \includegraphics[width=0.9\columnwidth]{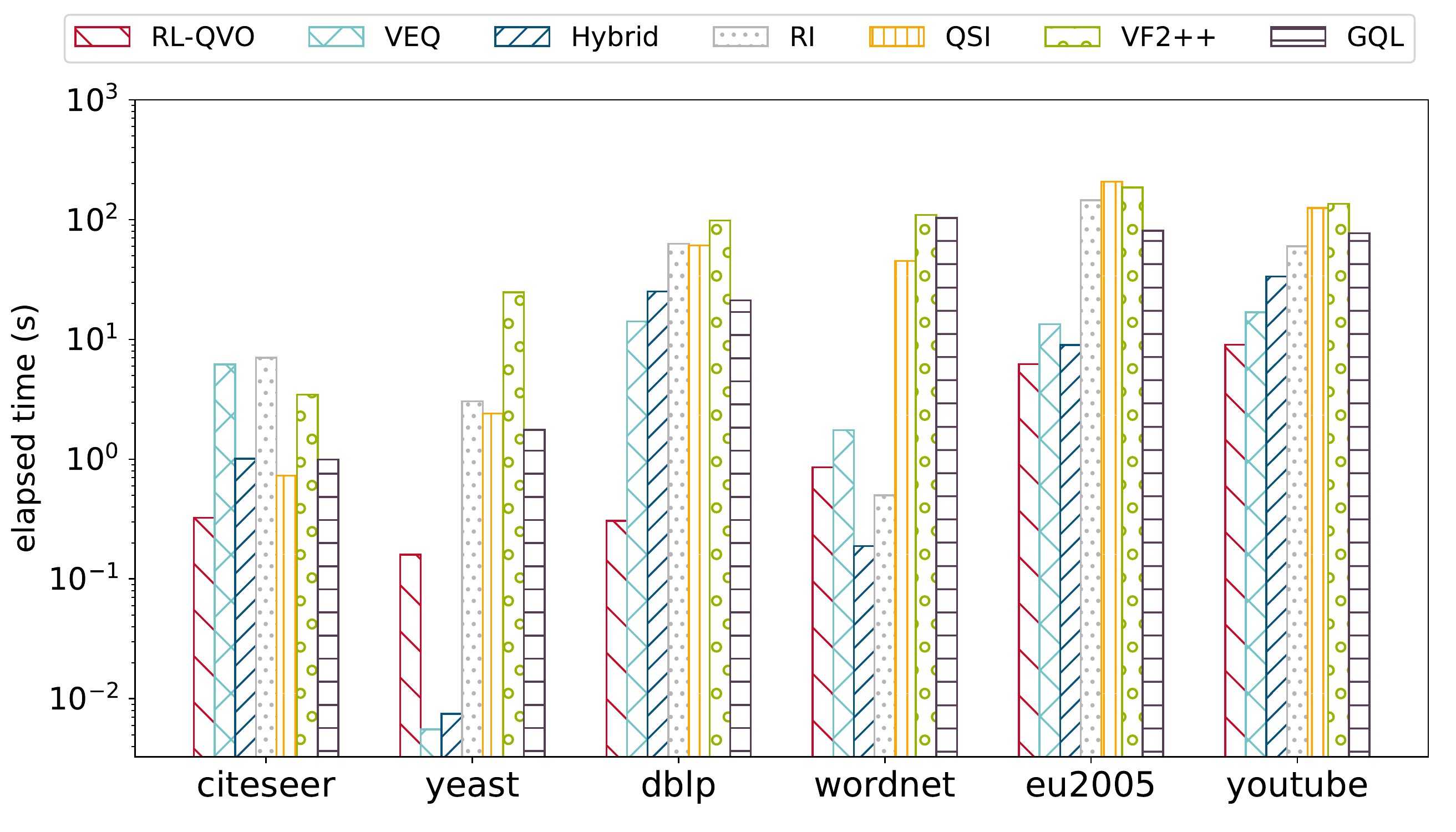}
    \vspace{-4mm}
    \caption{Average Query Processing Time Comparison}
    \vspace{-5mm}
    \label{fig:8_time_mean}
\end{figure}

\begin{figure}[tb]
    \centering
    \includegraphics[width=\columnwidth]{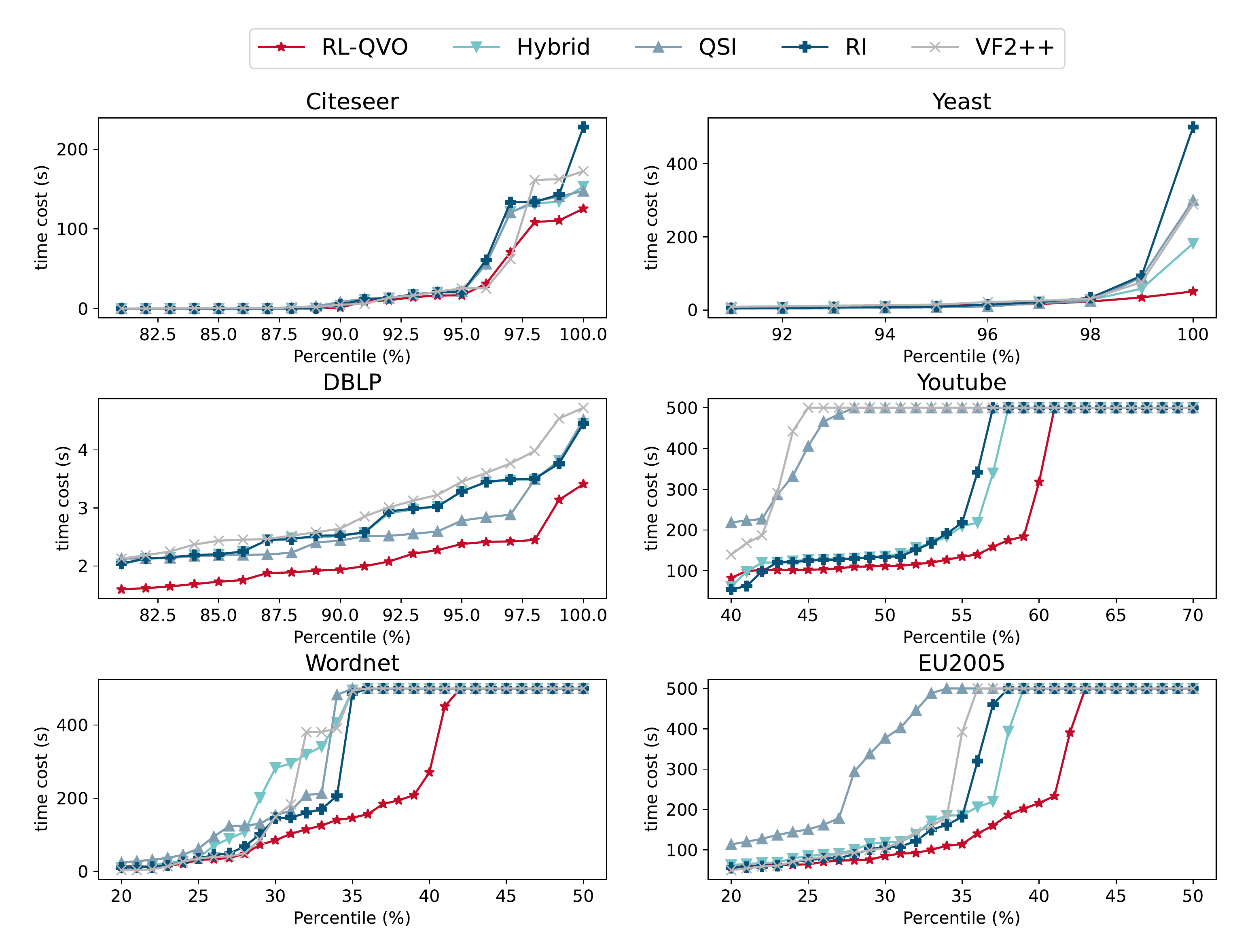}
    \vspace{-6mm}
    \caption{Query Processing Time Percentile Comparison}
    \vspace{-5mm}
    \label{fig:time_overall_percentile}
\end{figure}


\begin{figure}
\centering
\begin{subfigure}{.96\columnwidth}
  \centering
  \includegraphics[width=\linewidth]{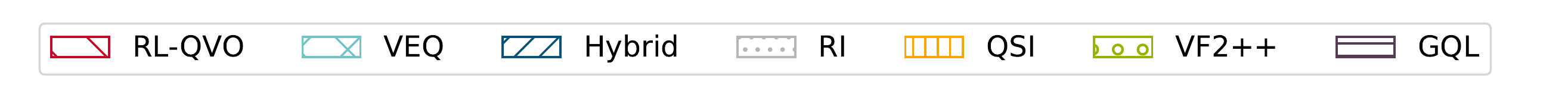}
\end{subfigure}%

\begin{subfigure}{.48\columnwidth}
  \centering
  \includegraphics[width=\linewidth]{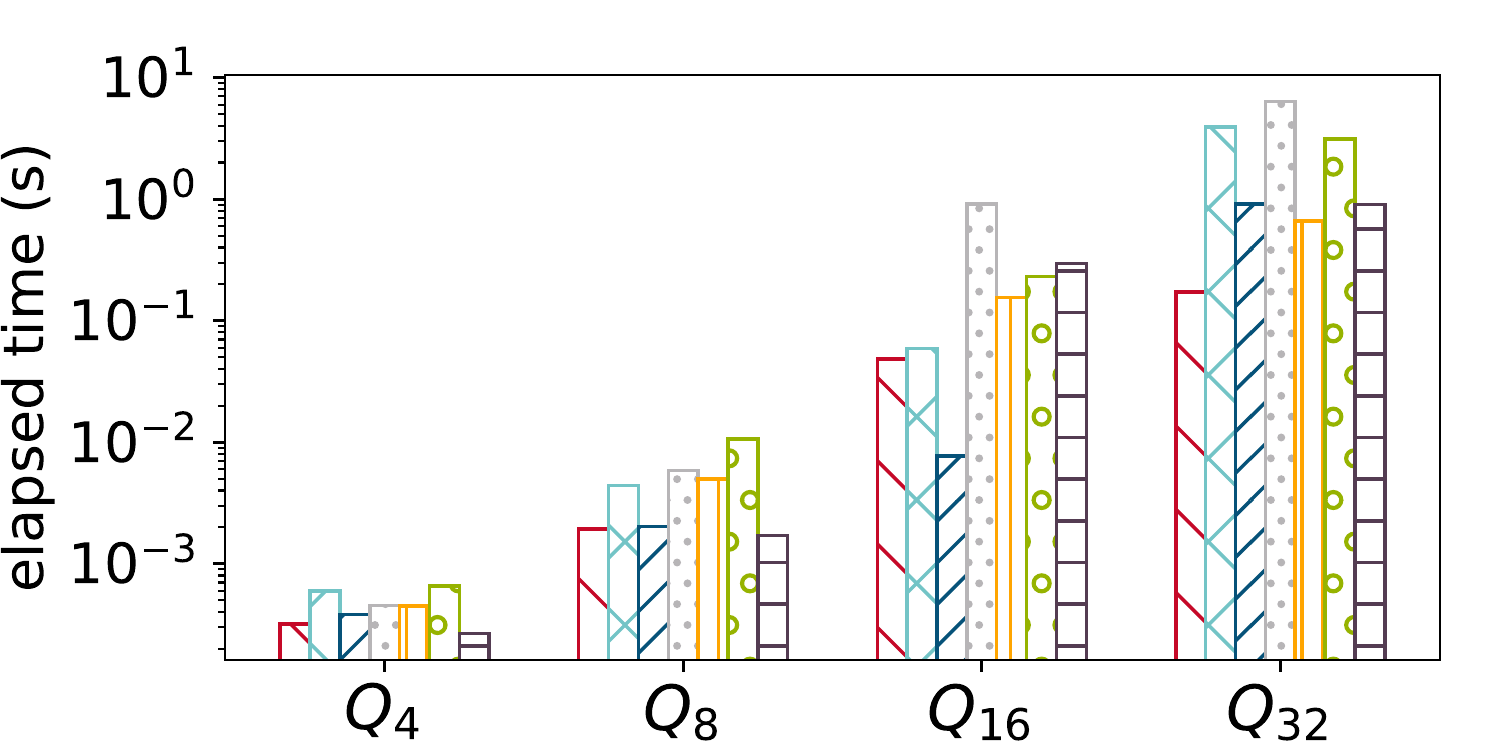}
  \vspace{-6mm}
  \caption{Citeseer}
  \label{fig:citeseer_enum}
\end{subfigure}%
\begin{subfigure}{.48\columnwidth}
  \centering
  \includegraphics[width=\linewidth]{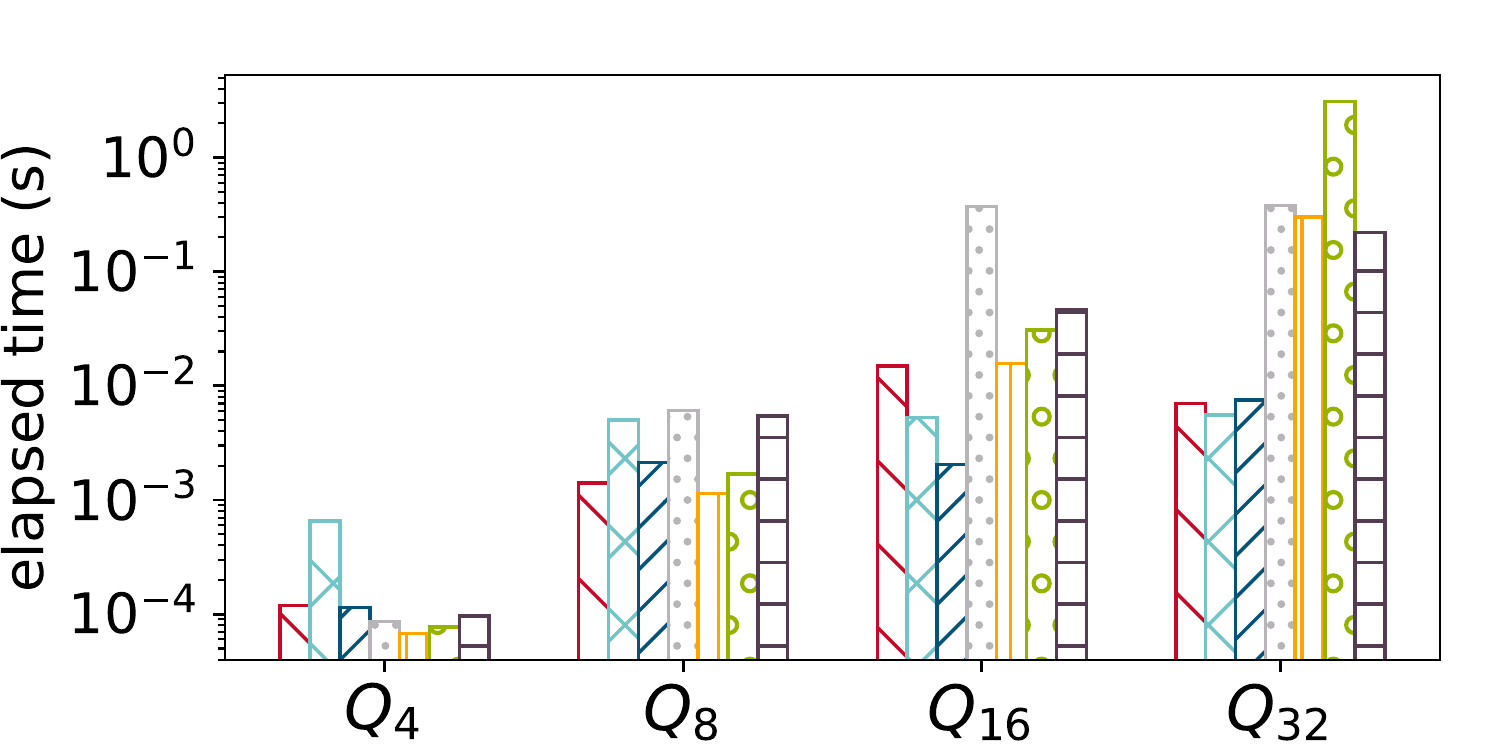}
  \vspace{-6mm}
  \caption{Yeast}
  \label{fig:yeast_enum}
\end{subfigure}
\begin{subfigure}{.48\columnwidth}
  \centering
  \includegraphics[width=\linewidth]{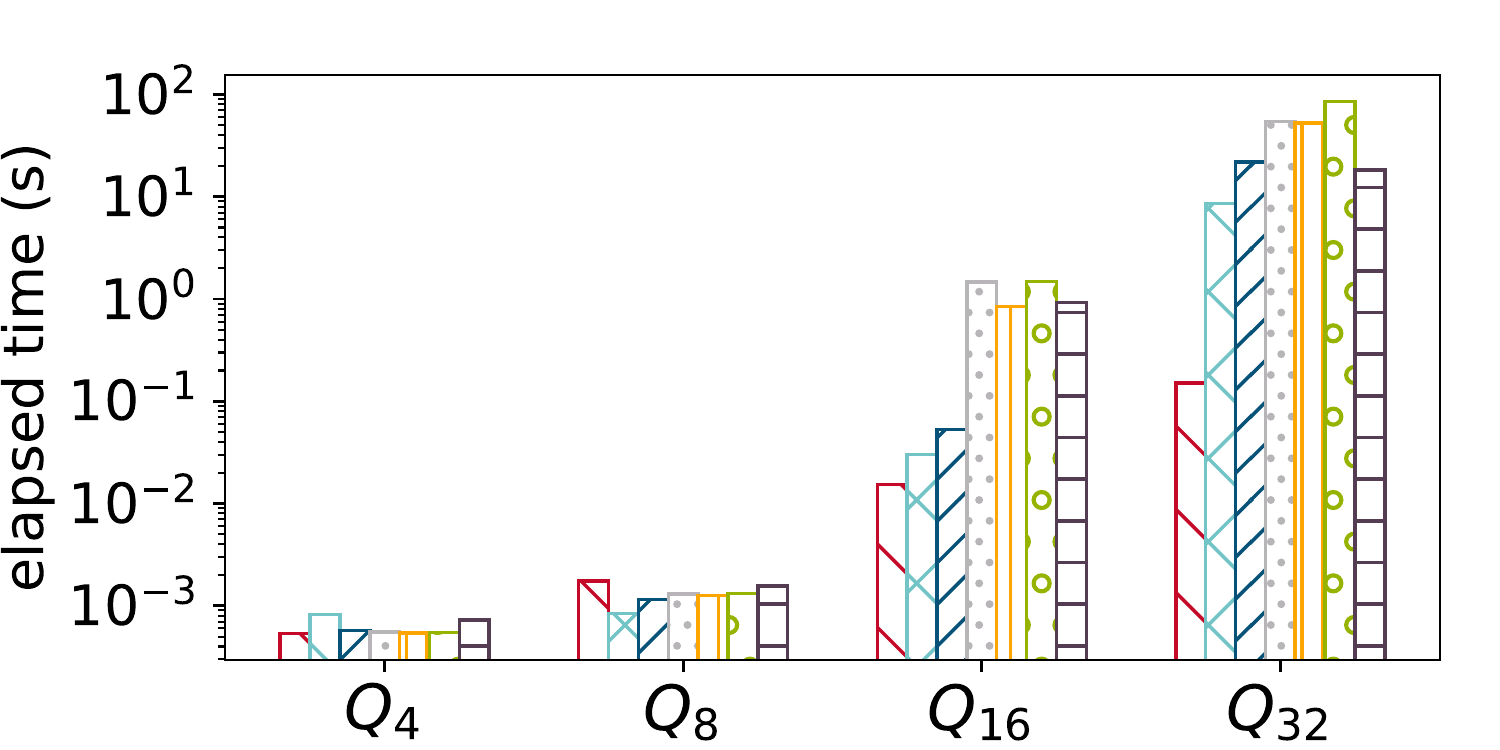}
  \vspace{-6mm}
  \caption{DBLP}
  \label{fig:dblp_enum}
\end{subfigure}
\begin{subfigure}{.48\columnwidth}
  \flushleft
  \includegraphics[width=\linewidth]{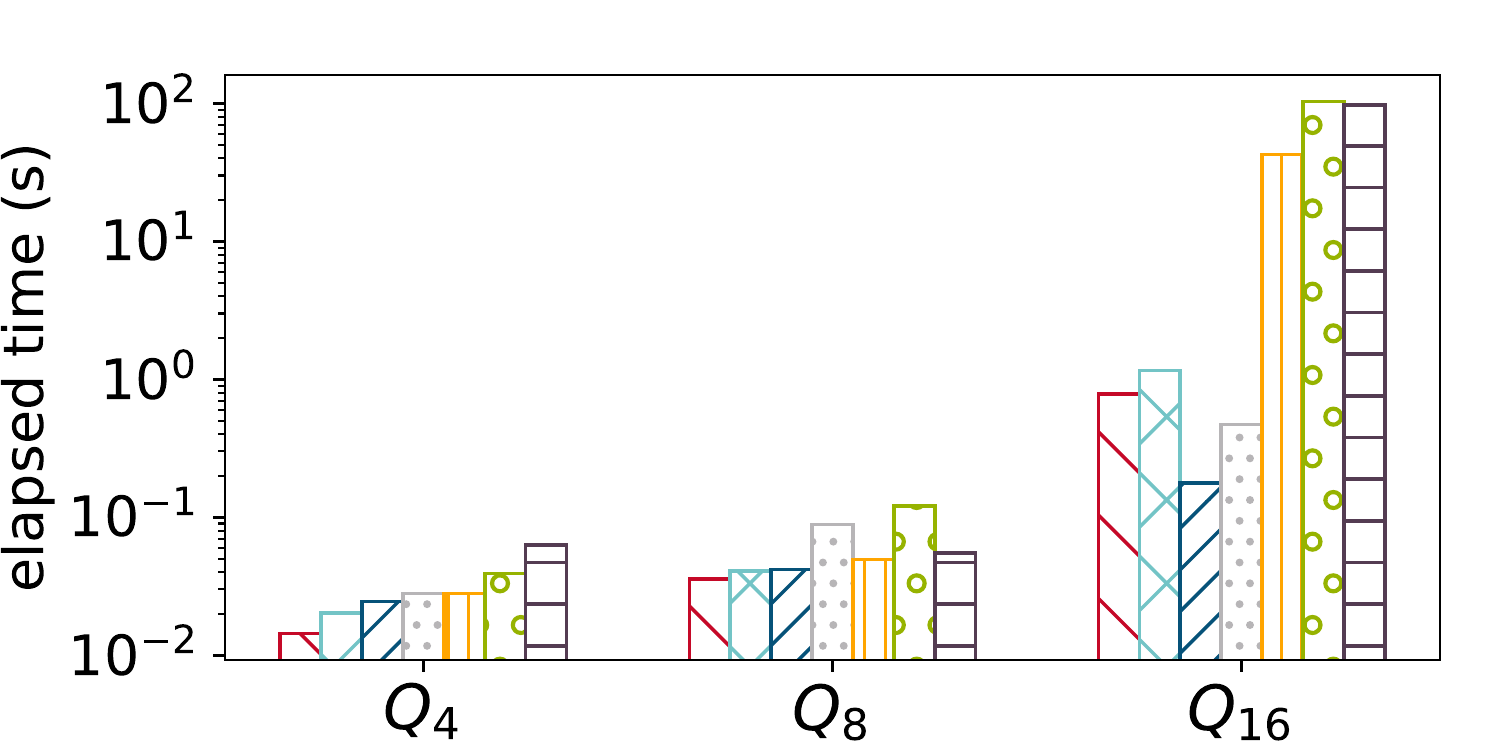}
  \vspace{-6mm}
  \caption{Wordnet}
  \label{fig:wordnet_enum}
\end{subfigure}
\newline
\begin{subfigure}{.48\columnwidth}
  \centering
  \includegraphics[width=\linewidth]{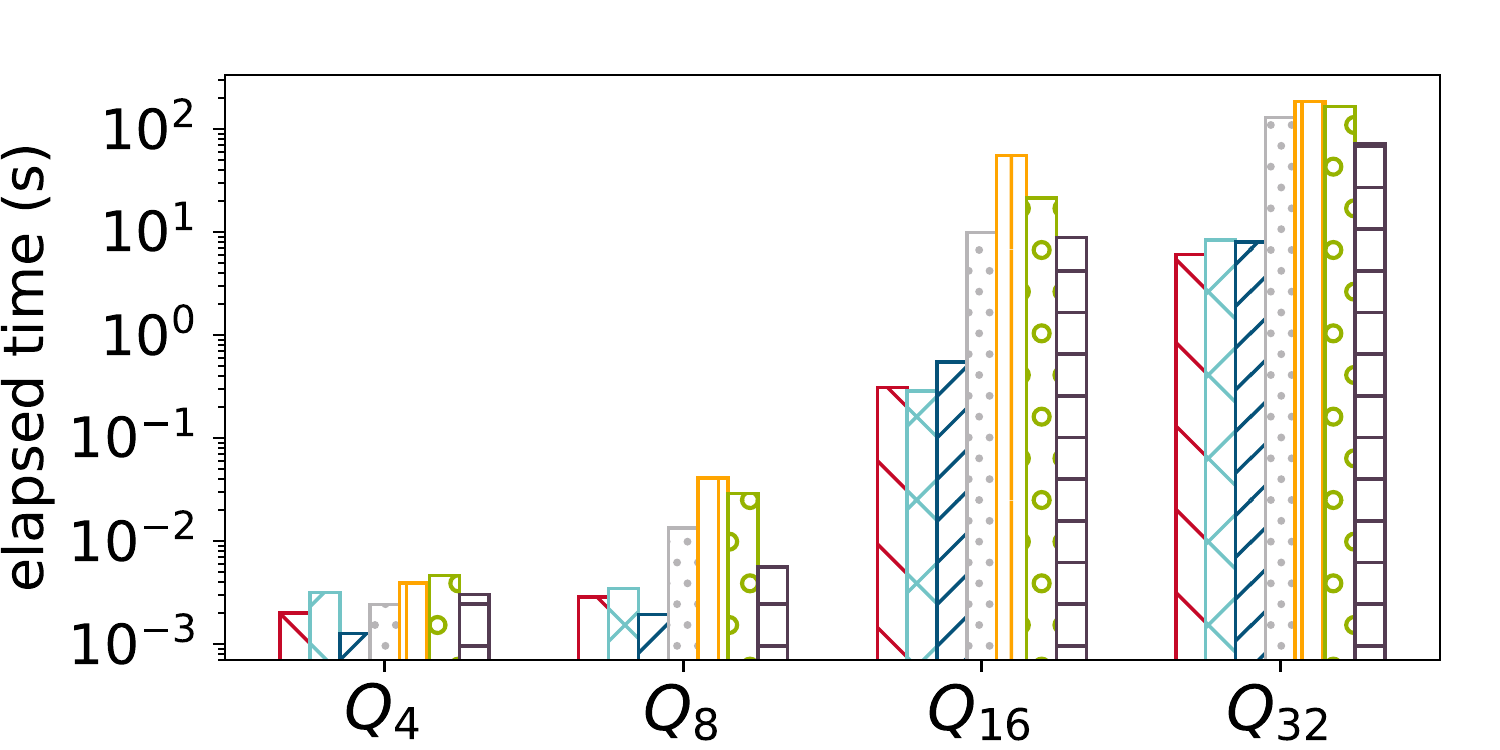}
  \vspace{-6mm}
  \caption{EU2005}
  \label{fig:eu2005_enum}
\end{subfigure}
\begin{subfigure}{.48\columnwidth}
  \centering
  \includegraphics[width=\linewidth]{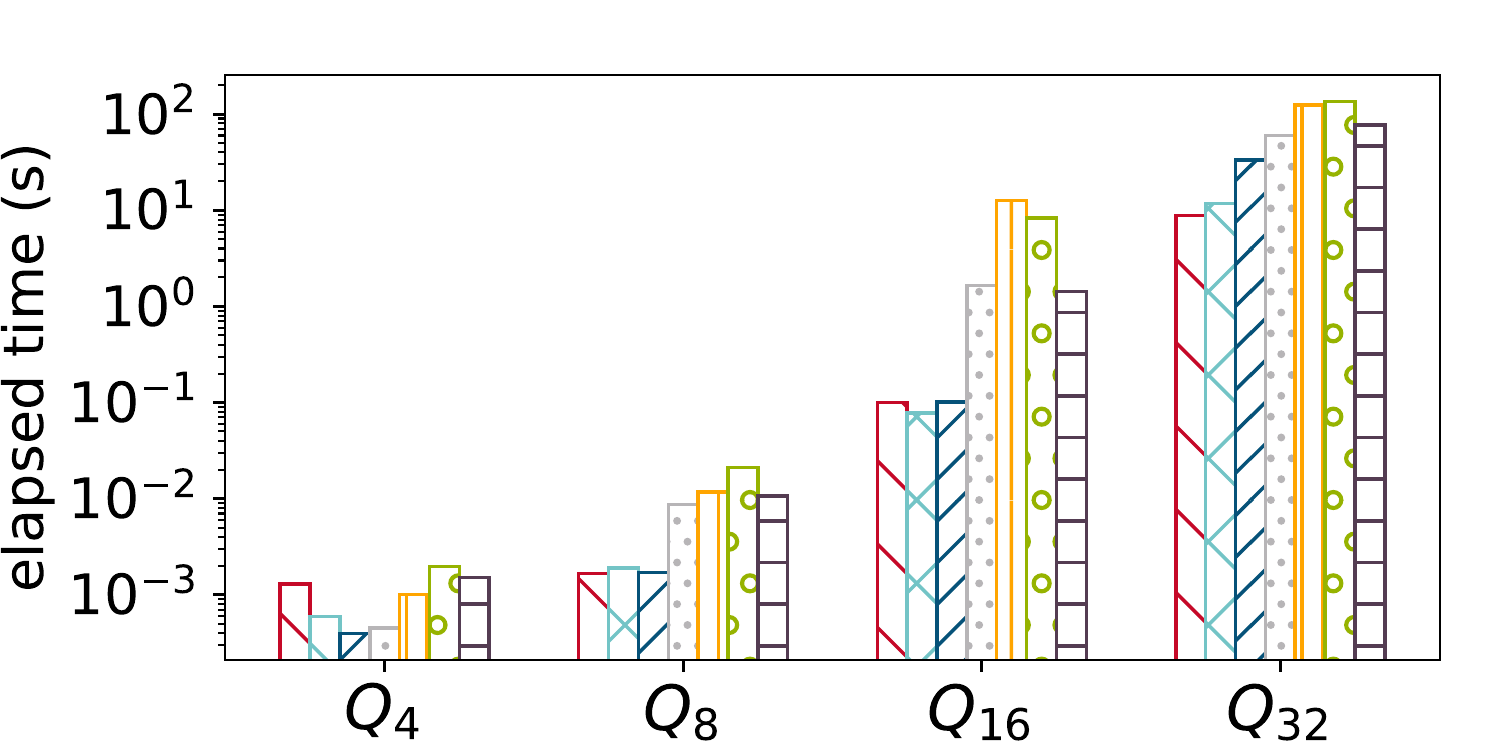}
  \vspace{-6mm}
  \caption{Youtube}
  \label{fig:youtube_enum}
\end{subfigure}
\caption{Enumeration time cost comparison}
\vspace{-5mm}
\label{fig:enumeration_all}
\end{figure}


\begin{figure}[tbh]
    \centering
    \includegraphics[width=0.9\columnwidth]{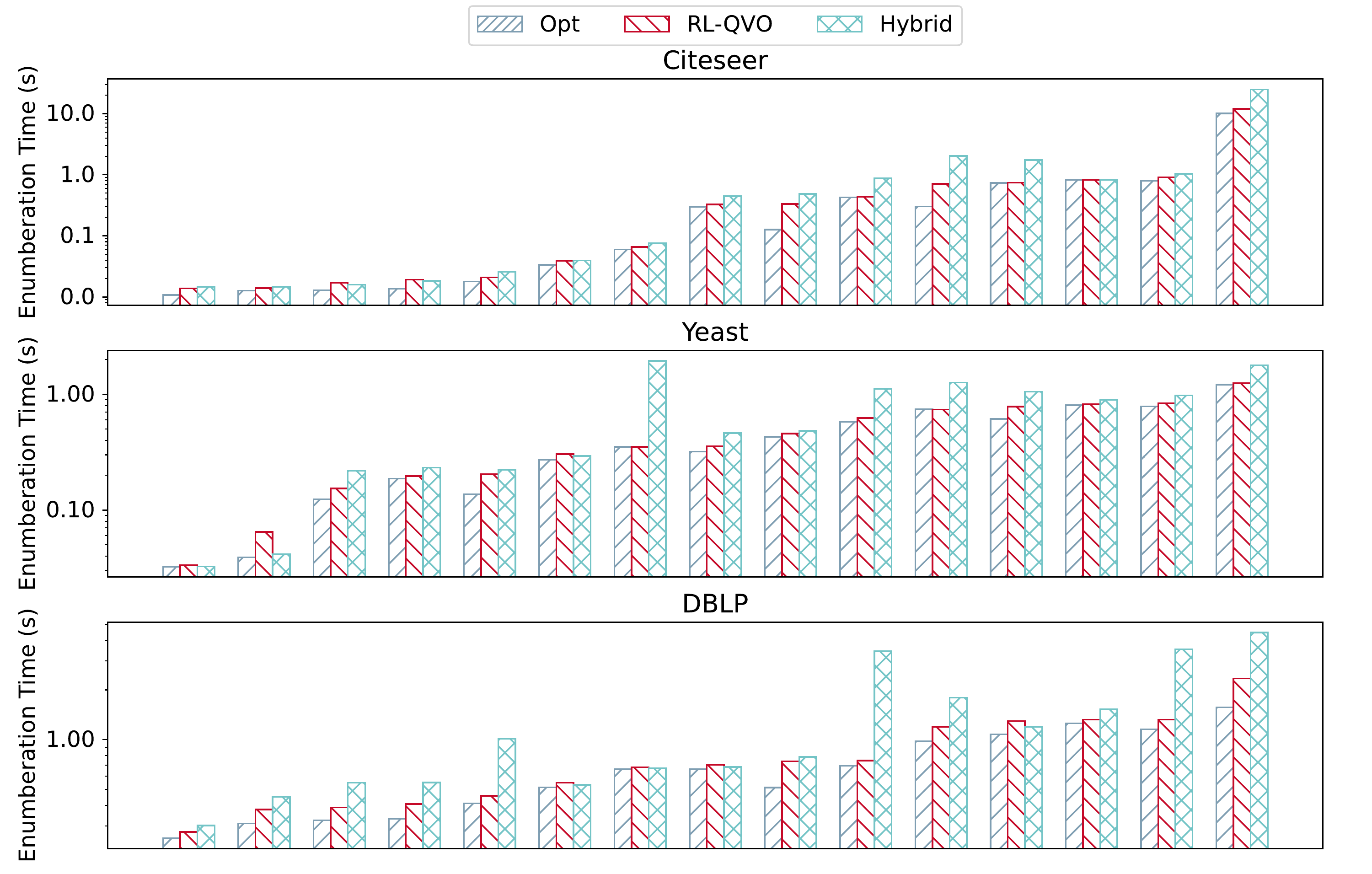}
    \vspace{-3mm}
    \caption{Enumeration time spectrum analysis\\ with optimal orders}
    \vspace{-7mm}
    \label{fig:enum_perm}
\end{figure}

In the first set of experiments, we evaluate the  query processing time for all compared subgraph matching algorithms
with default query graph sets, where $6$ real-life graphs are deployed. 
Note that the query processing time $t$ includes the filtering methods time cost $t_{filter}$, the ordering time $t_{order}$ and enumeration time $t_{enum}$, i.e., $t = t_{filter} + t_{order} + t_{enum}$.

\noindent \textbf{Average Query Processing Time.}
We first report the average query processing time for the given query graphs on default query sets, and the results are illustrated in Figure~\ref{fig:8_time_mean}.
It is shown that \mname\ generally outperforms the competitors because of the high quality matching order obtained by 
our advanced model. 
In particular, \mname outperforms VEQ by up to two orders of magnitude on Citeseer and DBLP.
\mname is also more than two orders of magnitude faster than \hname on DBLP.
In the experiments of Section~\ref{subsec:exp_enumeration_time}, we further justify that this is a significant achievement where the optimal matching order is considered.



\vspace{1mm}
\noindent \textbf{Cumulative Query Processing Time Distribution.} 
To better understand the query processing time distribution,
Fig.~\ref{fig:time_overall_percentile} details the query processing time comparison by showing a cumulative distribution of the query processing time for all compared methods.
Please note that in order to demonstrate the differences much clearer, the query processing time report in this experiment is the time cost to find \textit{all} subgraph matches in the corresponding data graphs, which is much greater than the time cost to find first $10^5$ matches.
The gaps of time cost between \mname and other compared methods grow with percentile, which shows the efficiency of \mname\ in handling the hard queries. This is a big advantage of \mname compared to other competitors because 
the response time of queries at high percentile (\textit{i.e.,} hard queries) is critical for the through-output of the system in many industry applications.

\vspace{1mm}
\noindent \textbf{Number of Unsolved Queries.}
Fig.~\ref{fig:time_overall_percentile} also demonstrates the number of unsolved query graphs with default query sets for Youtube, Wordnet and EU2005 data graphs. Note that we set the query time of each unsolved query to $500$ seconds (preset time limit). It is shown that \mname\ has much less number of unsolved queries compare to other algorithms.

\subsection{Enumeration Time Comparison}
\label{subsec:exp_enumeration_time}

The enumeration time is the determining factor of the overall time cost of subgraph matching process, and also directly reflects the qualities of matching orders generated by different algorithms.
We compare the enumeration time of VEQ, Hybrid, QSI, RI, VF2++ and GQL with \mname\ to examine the qualities of the matching orders generated by these methods.
Since all these methods utilize the same enumeration methods which are implemented in the same way, the enumeration time costs could directly reflect the qualities of the output matching orders.

\vspace{1mm}
\noindent \textbf{Average Enumeration Time.}
The average enumeration time with varying query sizes are shown in Fig.~\ref{fig:enumeration_all}.
It is reported that our proposed \mname has the state-of-the-art performance on all datasets with all query sizes.
Similar as query processing time, our proposed \mname\ could improve the enumeration time up to 2 orders of magnitude compared to the \textit{best} performed baseline method on $Q_{32}$ of DBLP. The performance gaps between \mname\ and other baseline methods become much more significant with the growth of query vertex numbers, which indicates that \mname\ has formidable ability in handling ordering problem in large search space.
It is also worth noting that \mname achieves similar performance as baseline algorithms w.r.t the enumeration time on Yeast data graph, which is much better than the time cost for query processing.
This result indicates that \mname might need relatively more time to generate matching order, but the generated order has high quality even on small data graphs like Yeast.

\vspace{1mm}
\noindent \textbf{Comparison with Optimal Matching Order.}
To better investigate the goodness and the potential improvement space of the matching orders obtained by 
the algorithms, we also consider the optimal matching orders in the experiments.
As it is time prohibitive to find the optimal matching orders for large size queries and data graphs.
We only consider the settings where there are no unsolved queries for \mname and \hname. 
Particularly, we conduct the spectrum analysis on Citeseer, Yeast and DBLP datasets to find \textit{all} subgraph matches with 15 randomly selected query graphs with $8$ vertices ($Q_8$) for each dataset.
To obtain the \textit{optimal matching order},
we generate the orders of all permutations of the query vertices, and feed them into the subgraph matching algorithm with the same filtering and enumeration methods as \mname\ and Hybrid.
We pick the permutation that requires the minimum enumeration number as the \textit{optimal matching order}.

We compare the enumeration time of \mname\ and \hname to the optimal matching order based 
algorithm, denoted by \textbf{Opt}. Note that all three algorithms use the same filtering and enumeration implementations.
The analysis results are illustrated in Fig.~\ref{fig:enum_perm} where the bars indicate the enumeration time costs for the optimal order, \mname's order and Hybrid's order. 
The results show that compared to the ordering methods of Hybrid, \mname\ is more likely to generate \textit{near optimal} orders for matching, thus have better overall search performance.
Moreover, considering the gaps between \textit{Opt} and \hname for these queries in Fig.~\ref{fig:enum_perm}, 
we boast that that \mname makes a significant improvement in terms of enumeration time (\textit{i.e.,} the quality of the matching order).

\subsection{Ablation Study}
\label{subsec:ablation}

\begin{figure}
\begin{subfigure}{\columnwidth}
  \centering
  \includegraphics[width=\linewidth]{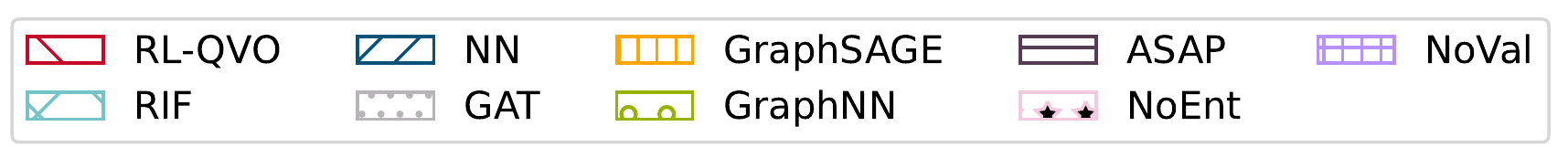}
  \vspace{-4mm}
  \label{fig:sfig0}
\end{subfigure}
\begin{subfigure}{.5\columnwidth}
  \centering
  \includegraphics[width=\linewidth]{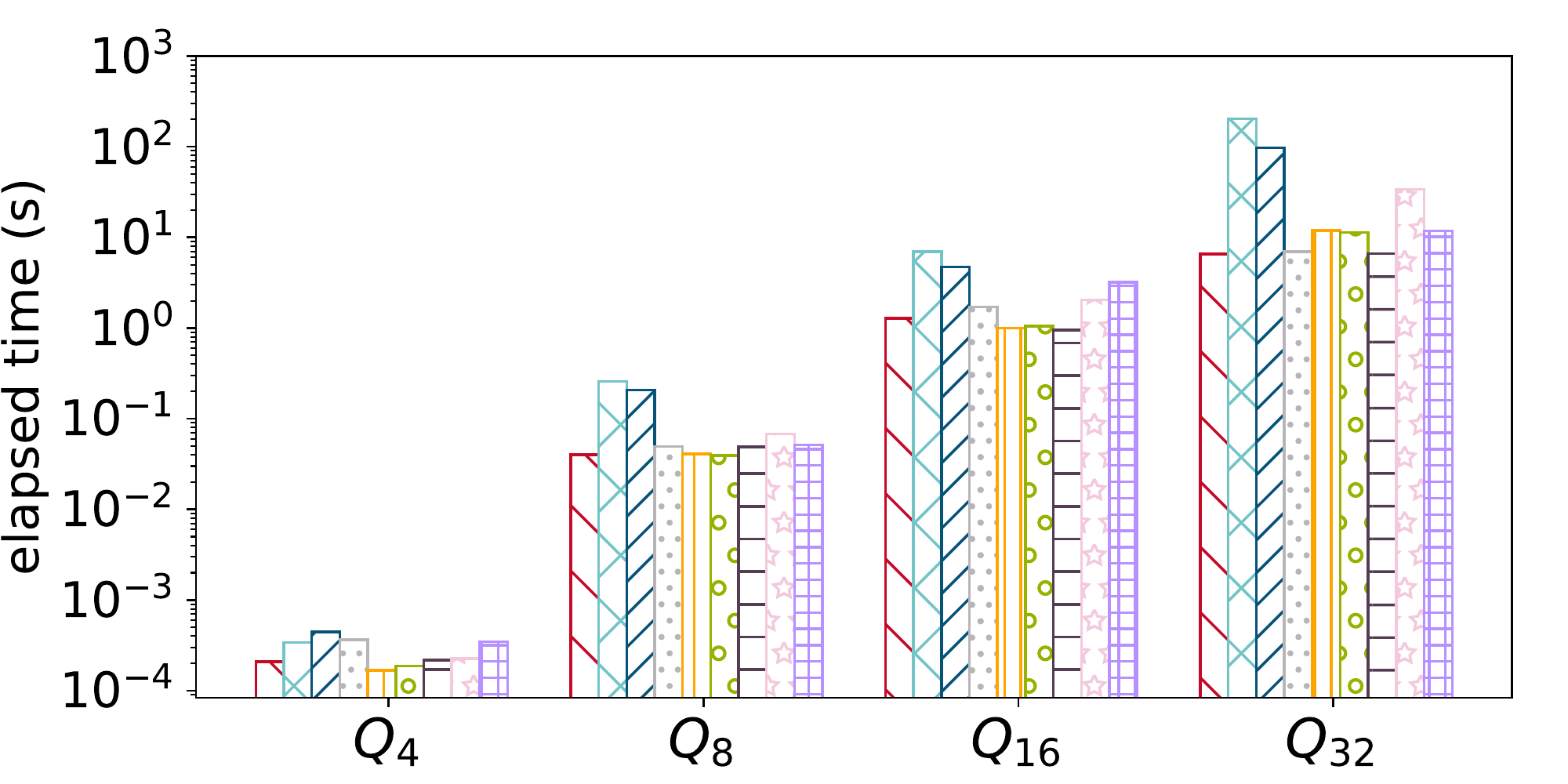}
  \vspace{-6mm}
  \caption{Query Processing Time}
  \label{fig:sfig1}
\end{subfigure}%
\begin{subfigure}{.5\columnwidth}
  \centering
  \includegraphics[width=\linewidth]{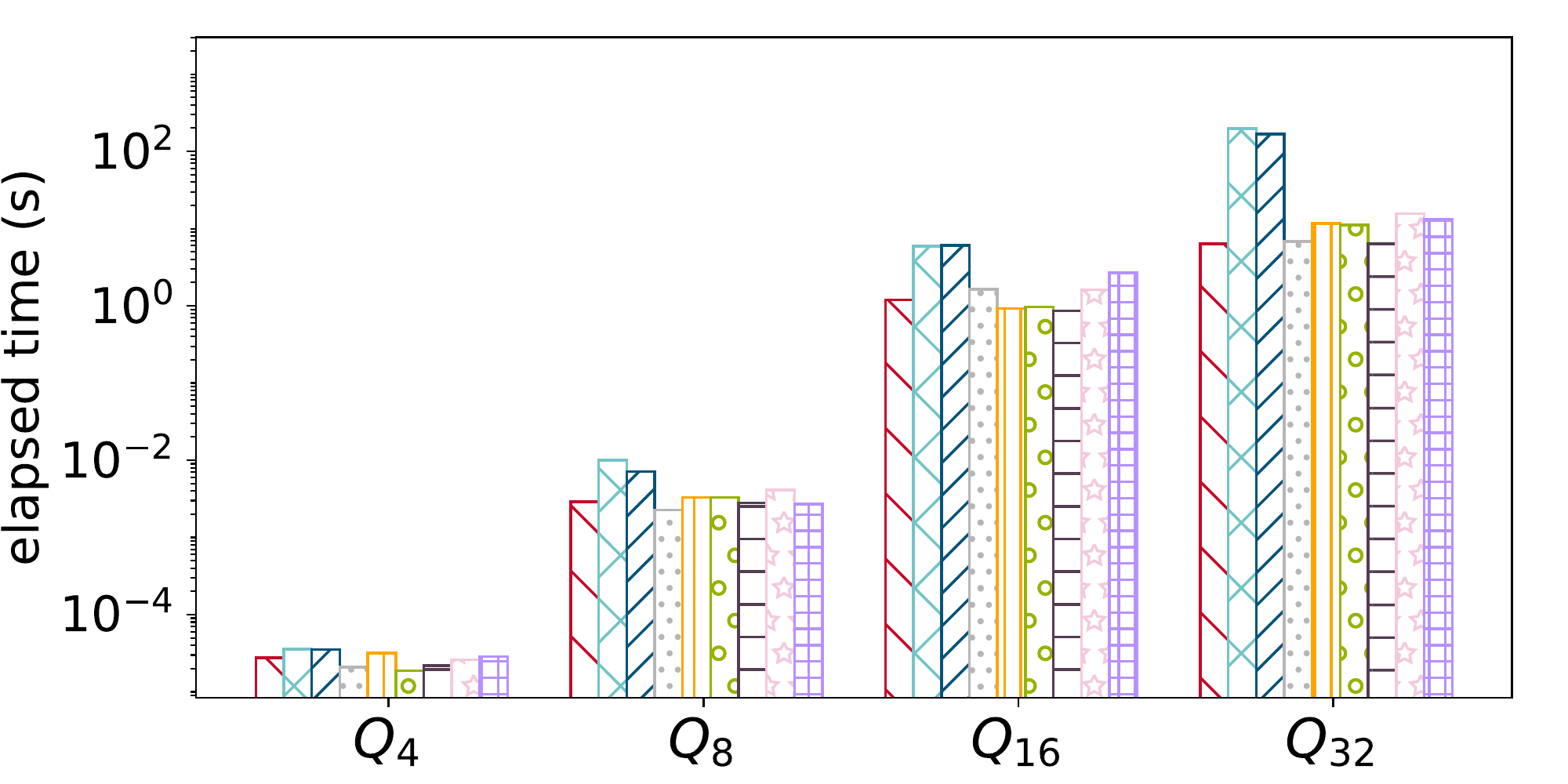}
  \vspace{-6mm}
  \caption{Enumeration Time}
  \label{fig:sfig2}
\end{subfigure}
\caption{Time comparison for ablation study on EU2005.}
\vspace{-5mm}
\label{fig:ablation_study}
\end{figure}

We evaluate the effectiveness and efficiency of different graph neural networks in this ablation study.
We change the agent components of \mname to get the following variants:

\noindent \textbf{RL-QVO-RIF} is the variant that each vertex has random input feature rather than our carefully designed features.

\noindent \textbf{RL-QVO-NN} is the variant that replaces the GCN with naive multi-layer perceptron (MLP) as the policy network, which tests the effectiveness of the GCN in this ordering task.

\noindent \textbf{RL-QVO-GAT} is the variant that uses graph attention network~\cite{DBLP:conf/iclr/VelickovicCCRLB18} as the policy network.

\noindent \textbf{RL-QVO-GraphSAGE} is the variant that uses GraphSAGE~\cite{hamilton2017inductive} as the policy network.

\noindent \textbf{RL-QVO-GraphNN} uses the GNN operator from~\cite{morris2019weisfeiler} as the policy network.

\noindent \textbf{RL-QVO-ASAP} exploits the GNN operator in ASAP~\cite{ranjan2020asap} as the policy network.

\noindent \textbf{RL-QVO-NoEnt} is the variant without the entropy reward.

\noindent \textbf{RL-QVO-NoVal} is the variant without the validation reward.

In this experiment, we compare the query processing and enumeration time of \mname and its variants on data graph EU2005 with query sets $Q_4$, $Q_8$, $Q_{16}$ and $Q_{32}$.
The results are illustrated in Fig.~\ref{fig:ablation_study}. 
Specifically, the significant performance difference between \mname and RL-QVO-NN validates the effectiveness of graph neural network in our model.
It can also be concluded from the results that different GNN models has limited differences w.r.t the enumeration time, i.e., the performance of \mname is not bound to the selection of GNN model.
The time cost gap between RL-QVO-RIF and \mname proves the effectiveness of our carefully designed feature.
Furthermore, the time difference between RL-QVO-NoEnt/NoVal and \mname shows the necessity of the entropy and validate rewards, especially with the large-size query sets.

\subsection{Query Processing Time vs. Output Dimension}

\begin{figure}
\centering
\begin{minipage}{.5\columnwidth}
  \centering
  \includegraphics[width=\linewidth]{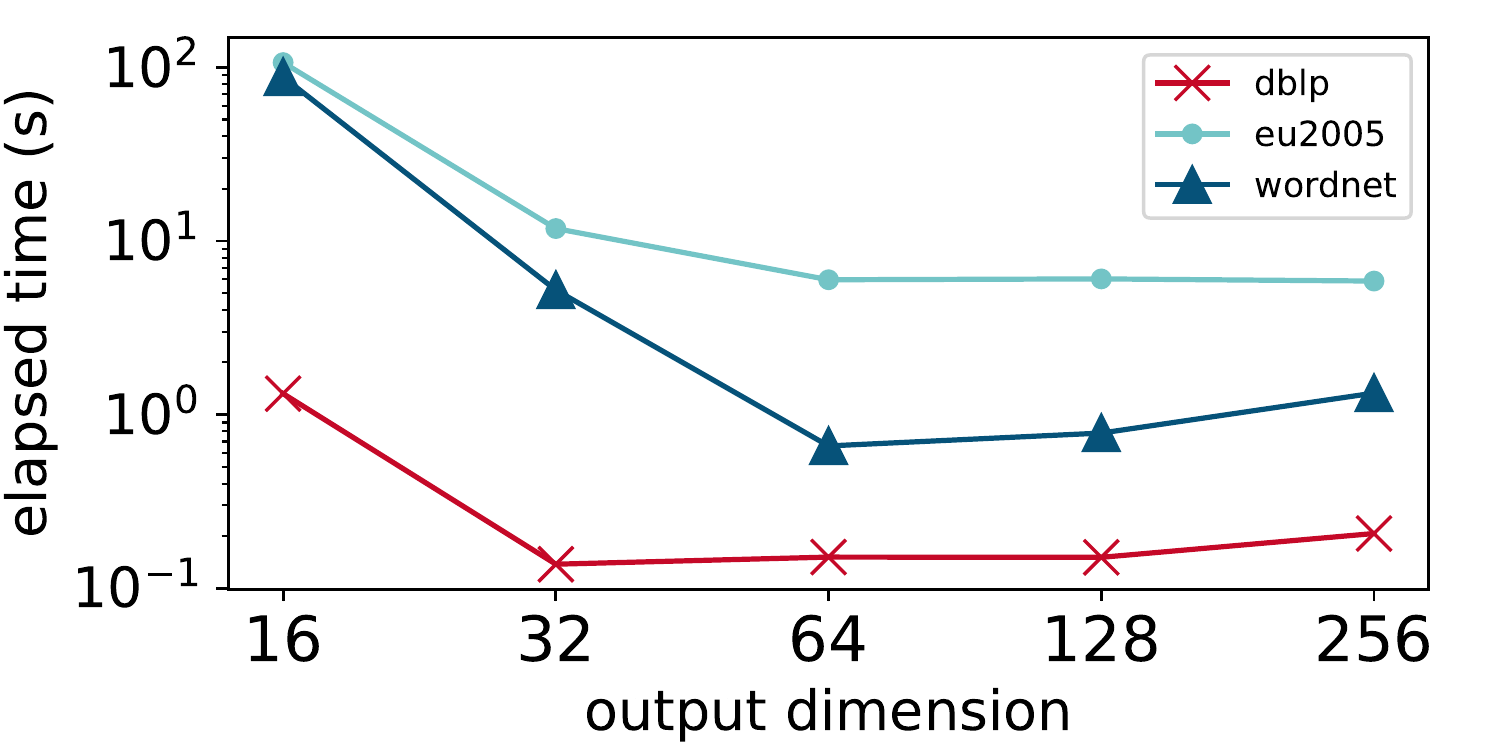}
  \captionof{figure}{Query processing time vs. output dimension}
  \label{fig:dimension}
\end{minipage}%
\begin{minipage}{.5\columnwidth}
  \centering
  \includegraphics[width=\linewidth]{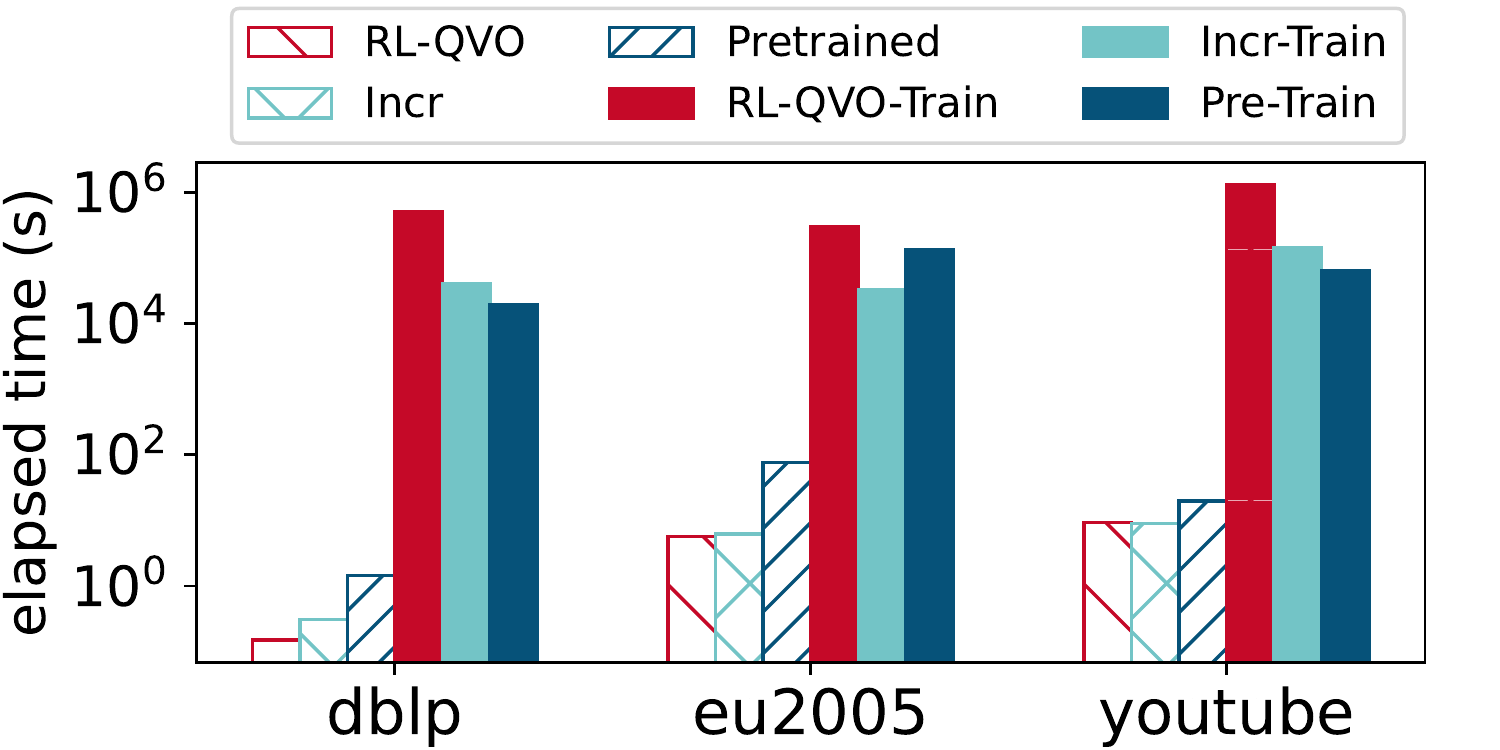}
  \captionof{figure}{Query processing time and training time comparison of incremental training.}
  \label{fig:incremental}
\end{minipage}
\vspace{-5mm}
\end{figure}

\begin{figure}
\centering
\begin{minipage}{.5\columnwidth}
  \centering
  \includegraphics[width=\linewidth]{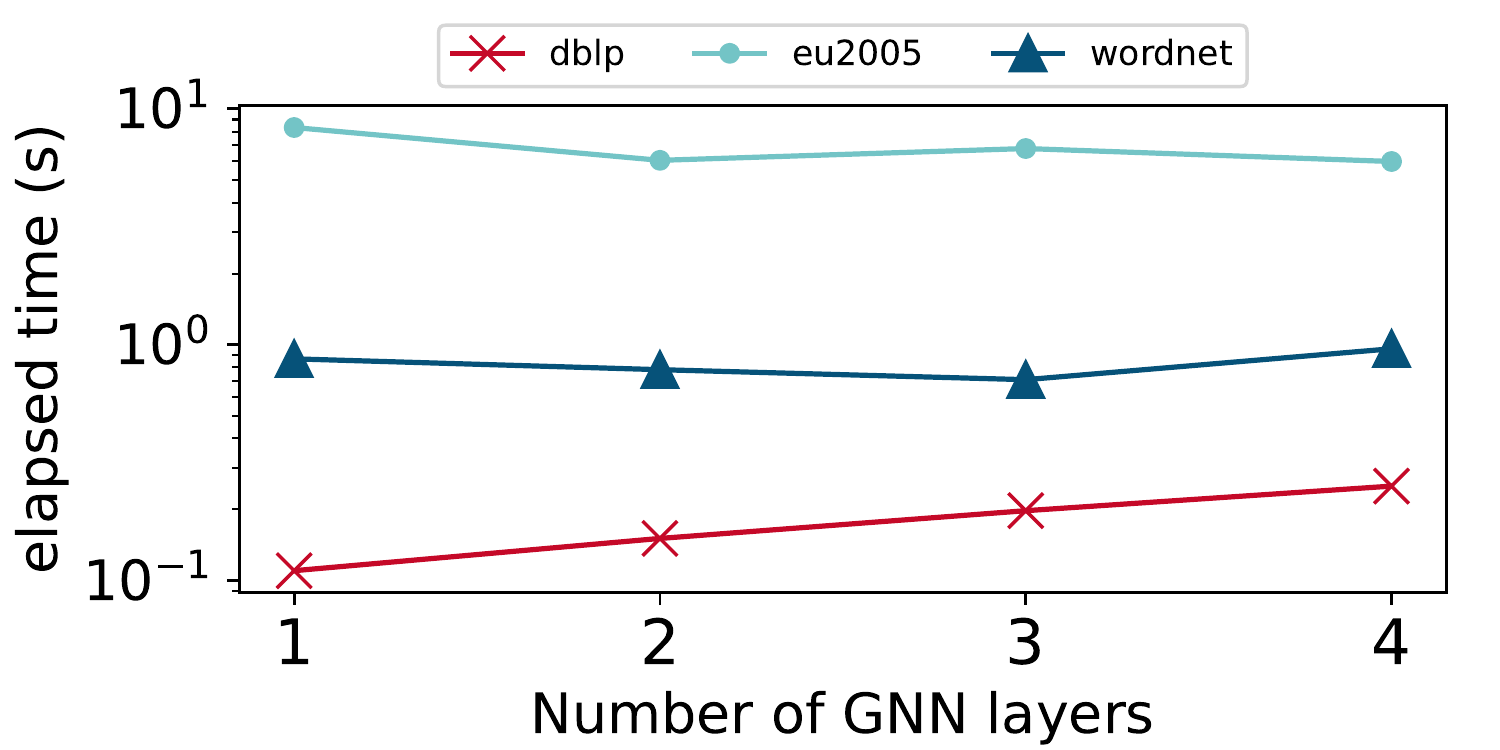}
  \captionof{figure}{Query processing time vs. number of GNN layers}
  \label{fig:num_layers}
\end{minipage}%
\begin{minipage}{.5\columnwidth}
  \centering
  \includegraphics[width=\linewidth]{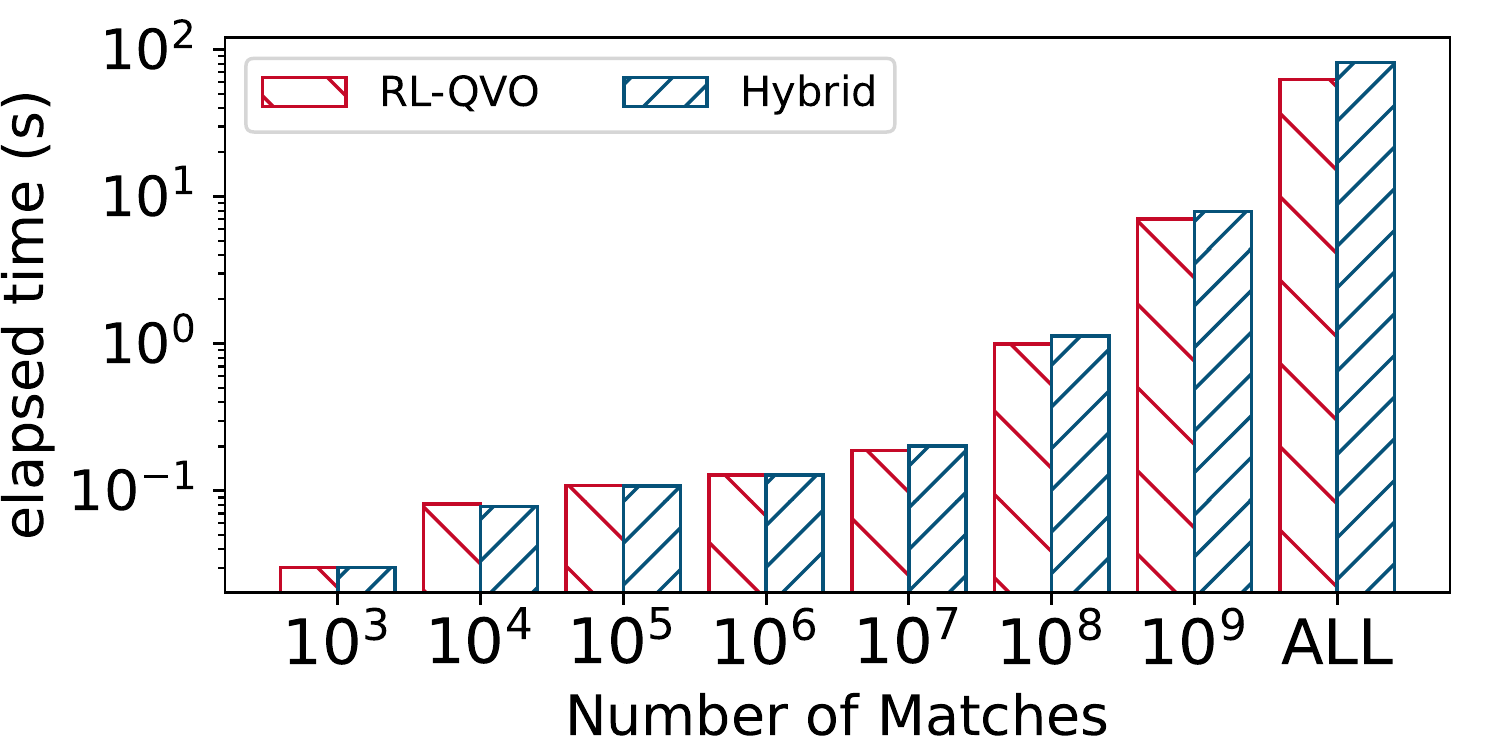}
  \captionof{figure}{Enumeration time vs. number of matches.}
  \label{fig:matches}
\end{minipage}
\vspace{-5mm}
\end{figure}



In this subsection, we analyze how the output dimension of \mname influence the query processing time.
We vary the output dimension of \mname in $\{16, 32, 64, 128, 256\}$ and conduct the experiment on DBLP, EU2005 and Wordnet with their default query sets.
The results are summarized in Fig.~\ref{fig:dimension}.
When the output dimension is $32$, \mname has limited performance because of the small parameter space.
With the growth of output dimension, the query processing time reduces, and the salient point is around 64.
When the dimension continues growing, the query processing time will increase because the time cost for ordering $t_{order}$ increases with the growth of output dimension.

\subsection{Training Time and Order Inference Time}
\label{subsec:incre_exp}


In this subsection, we evaluate the query processing time and training time of \mname to show the efficiency brought by the incremental training.

As mentioned in Section~\ref{sec:incr_train}, \mname could incorporate the incremental training to reduce the overall training time.
We compare three training method in this experiment.
(1) We train the model with the default query set for 100 epochs on corresponding data graphs.
(2) For each data graph, we first train the model on query set with less query vertices like $Q_8$ (for Wordnet) or $Q_{16}$ (for other data graphs) for 100 epochs.
Then incremental training is applied on default query set for 10 epochs.
(3) The pre-trained model is directly applied on default query set.
The results are illustrated in Fig.~\ref{fig:incremental}.
Model trained on default query set for 100 epochs (denoted as \mname in Fig.~\ref{fig:incremental}) always has the least query processing time.
Directly applying the pre-trained model (denoted as Pretrained) usually has significant performance gap with RL-QVO.
Compared with above training methods, the incremental trained model (denoted as Incr) could save nearly two orders of magnitude of training time while only sacrificing negligible performance.
We would like to emphasize that all the time cost results reported in previous experiments are produced by incremental training method.


\vspace{1mm}
\noindent \textbf{Order inference time.} 
As analyzed in Section~\ref{subsec:alg_complexity}, the computational complexity of \mname\ for query vertex ordering
generation is $O(\lvert V(q) \rvert \times (\lvert E(q) \rvert + d^2))$.
In our experiments, \mname\ could produce the matching order within $100ms$.


\subsection{Query Processing Time vs. Number of GNN layers}
We analyse how the number of GNN layers influences the query processing time in this subsection.
We vary the number of GNN layers in $\{1, 2, 3, 4\}$ and conduct the experiment on  on DBLP, EU2005 and Wordnet with their default query sets.
The results are illustrated in Fig.~\ref{fig:num_layers}.
When the data graph is relatively small, \textit{e.g.,} DBLP, the query processing time increases with the number of GNN layers near-linearly, since the query processing time is mainly consist of the time to generate the matching order.
Otherwise, on the lager data graphs, the model with only one GNN layer usually has the worst performance due to its limited power to leverage the structural information in the graphs.
When the number of layers is more than one, there is no significant difference between the query processing time, except the time cost for model with 4 layers on Wordnet.

\subsection{Enumeration Time vs. Number of Matches}
As mentioned in Section~\ref{sec:setup}, we measure the matching time to find the first $10^5$ matches in most of previous experiments.
In this subsection, we report the average enumeration time of \mname and \hname varying the number of matches enumerated on data graph Youtube with $Q_{16}$.
The results are shown in Fig.~\ref{fig:matches}, where \textit{ALL} denotes the experiment setting that finds all the matches of a query graph in the data graph.
Please note that the processing time of query that exceeds the time limit ($500$ secs) is set to 500 seconds, and the time costs of queries that cannot be processed by both methods within the time limit are not reported in this experiment.
As illustrated in Fig.~\ref{fig:matches}, when the number of matches is relatively small (from $10^3$ to $10^6$), there is no notable time difference between \mname and \hname.
With the growth of number of enumerated matches, \mname has more significant advancement compared to \hname in terms of the enumeration time, which indicates the superior order generation ability of \mname over \hname on large search space.

\subsection{Space Evaluation}

\begin{table}[tb]
\centering
\caption{Space Evaluation}
\vspace{-2mm}
\begin{tabular}{|c|c|c|}
\hline
\textbf{Dataset}  & \textbf{Graph Space} & \textbf{Model Space} \\ \hline
\textbf{Citeseer} & 112.4 kB             & 186.2 kB             \\ \hline
\textbf{Yeast}    & 260.8 kB             & 186.2 kB             \\ \hline
\textbf{DBLP}     & 30.4 MB              & 186.2 kB             \\ \hline
\textbf{Youtube}  & 89.7 MB              & 186.2 kB             \\ \hline
\textbf{Wordnet}  & 3.5 MB               & 186.2 kB             \\ \hline
\textbf{EU2005}   & 437.6 MB             & 186.2 kB             \\ \hline
\end{tabular}
\vspace{-4mm}
\label{tb:space}
\end{table}


As discussed in Section~\ref{subsec:alg_complexity}, the space complexity for \mname\ is fixed with growing sizes of query and data graphs.
Here, we report the exact space requirements for storing the parameters of the network and corresponding data graphs.
The results are demonstrated in Table~\ref{tb:space}.
Even for data graph like EU2005 which requires $437.6$ MB to store, \mname\ only needs $186.2$ kB to save parameters while achieving outstanding performance in query vertex ordering.
In terms of the memory consumption, \mname\ might require more space for query graphs, however, the memory cost is still dominated by the parameter space of the model, which is relatively small.


\section{Conclusion}
\label{sec:conclusion}
Subgraph matching is a fundamental research topic in database and data mining communities,
which is a NP-Complete problem.
One important branch of the subgraph matching algorithms follows the backtracking search paradigm,
and it is shown that the search performance is heavily affected by the quality of the matching order
used in the search. 
We notice that existing ordering methods with heuristic strategies are lacking the capability to fully exploit the structural and label information to generate high-quality matching order (i.e., query vertex order).
In this paper, we proposed \mname, a reinforcement learning-based model for query vertex ordering.
\mname\ learns the policy to generate the matching order considering both graph structure information
and the long-term benefits.
Extensive experiments prove the efficiency of \mname.

\newpage
\vspace{1mm}
\noindent \textbf{Acknowledgement.} Y. Zhang is supported by ARC DP210101393, L. Qin is supported by ARC FT200100787 and DP210101347, W. Wang was supported by HKUST(GZ) Grants G0101000028 and GZU22EG04, W. Zhang is supported by FT210100303 and DP200101116
\balance
\bibliographystyle{IEEEtran}
\bibliography{reference}
\end{document}